\pdfoutput=1
\documentclass[11pt]{article}
\usepackage{CJKutf8}

\usepackage[preprint]{acl}

\usepackage{times}
\usepackage{latexsym}

\usepackage[T1]{fontenc}
\usepackage[utf8]{inputenc}

\usepackage{microtype}

\usepackage{inconsolata}

\usepackage{booktabs}
\usepackage{multirow}
\usepackage{multicol}
\usepackage{graphicx}
\usepackage{ulem}
\usepackage{float}
\usepackage{amsmath}
\usepackage{amssymb}
\usepackage{amsfonts}
\usepackage{pifont}
\usepackage[ruled]{algorithm2e}
\usepackage{longtable}
\usepackage{hyperref}

\definecolor{mydarkgreen}{rgb}{0.2,0.7,0.2}
\title{Enabling Weak LLMs to Judge Response Reliability \\via Meta Ranking}

\author{Zijun Liu$^1$ \ 
Boqun Kou$^2$ \ 
Peng Li$^{3*}$ \ 
Ming Yan$^4$ \
Ji Zhang$^4$ \
Fei Huang$^4$ \
Yang Liu$^{1,3*}$ \\
$^1$ Department of Computer Science and Technology, Tsinghua University, Beijing, China \\
$^2$ Weiyang College, Tsinghua University, Beijing, China \\
$^3$ Institute for AI Industry Research (AIR), Tsinghua University, Beijing, China \\
$^4$ Institute of Intelligent Computing, Alibaba Group
}

\begin{document}
\maketitle

\begin{abstract}
{\let\thefootnote\relax\footnotetext{*\ 
Corresponding authors: P.Li (\href{mailto:pengli09@gmail.com}{\texttt{pengli09@gmail.com}}) and Y.Liu (\href{mailto:liuyang2011@tsinghua.edu.cn}{\texttt{liuyang2011@tsinghua.edu.cn}}) }}

Despite the strong performance of large language models (LLMs) across a wide range of tasks, they still have reliability issues. Previous studies indicate that strong LLMs like GPT-4-turbo excel in evaluating the reliability of responses from LLMs, but face efficiency and local deployment issues. Thus, to enable weak LLMs to effectively assess the reliability of LLM responses, we propose a novel cross-query-comparison-based method called \emph{Meta Ranking} (MR). Unlike previous few-shot methods that solely based on in-context learning capabilities in LLMs, MR assesses reliability by pairwisely ranking the target query-response pair with multiple reference query-response pairs. We found that MR is highly effective in error detection for LLM responses, where weak LLMs, such as Phi-2, could surpass strong baselines like GPT-3.5-turbo, requiring only five reference samples and significantly improving efficiency.
We further demonstrate that MR can enhance strong LLMs' performance in two practical applications: model cascading and instruction tuning. In model cascading, we combine open- and closed-source LLMs to achieve performance comparable to GPT-4-turbo with lower costs. In instruction tuning, we use MR for iterative training data filtering, significantly reducing data processing time and enabling LLaMA-7B and Phi-2 to surpass Alpaca-13B with fewer training tokens. These results underscore the high potential of MR in both efficiency and effectiveness.\footnote{The source code is available at \url{https://github.com/THUNLP-MT/MetaRanking}.}

\end{abstract}

\begin{figure*}[t]
    \centering
    \includegraphics[width=\linewidth]{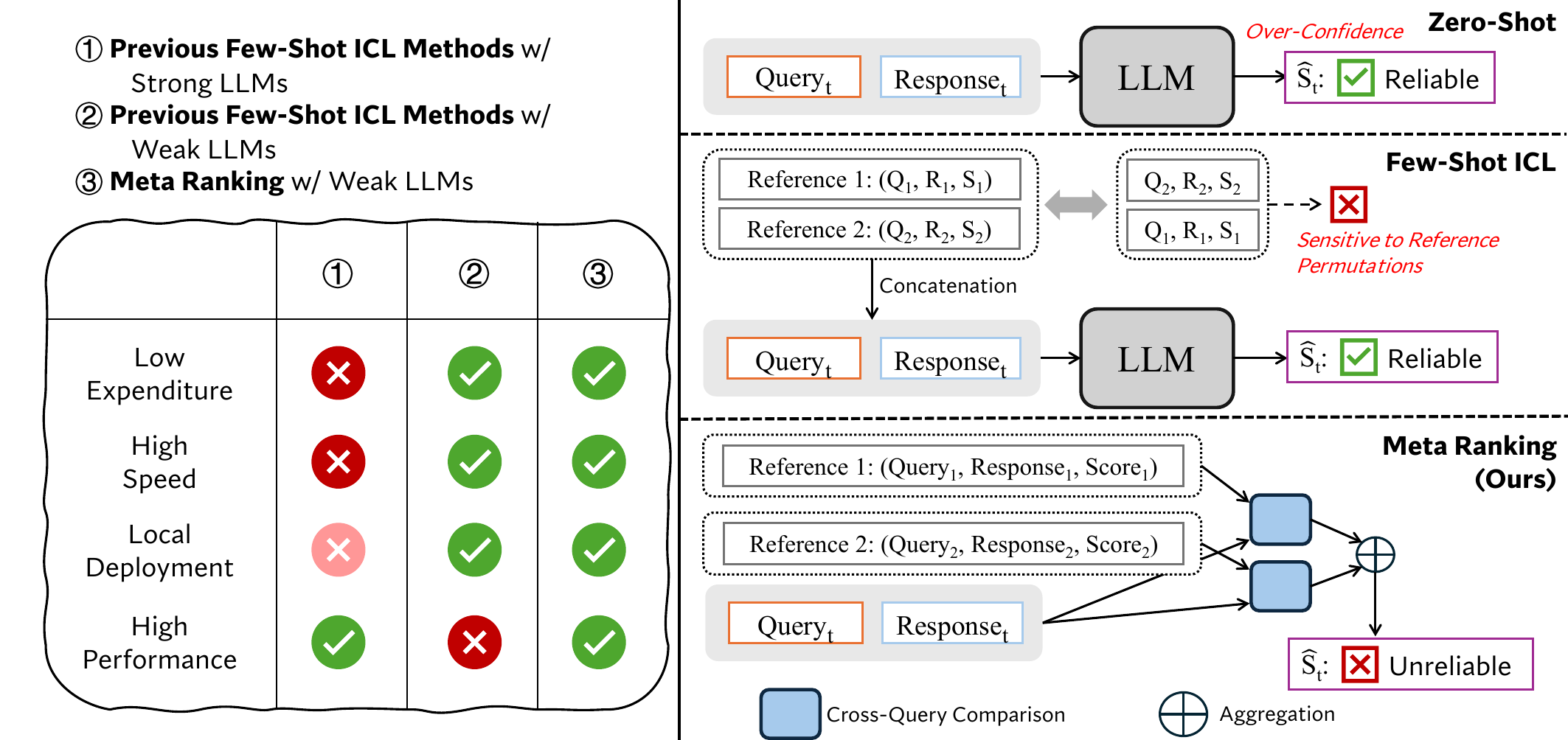}
    \caption{Overview of our proposed \emph{Meta Ranking} (MR) method. (a) Left: The table summarizes MR and previous judgement methods with different backbone models. (b) Right: The sub-figure illustrates different methods. ``$\widehat{\text{S}}_\text{t}$'' denotes the estimated score for the target query-response pair. ``Query$_i$'' (Q$_i$), ``Response$_i$'' (R$_i$), and ``Score$_i$'' (S$_i$) ($i=1,2$) denote the references and its score (e.g., +1 for correct and -1 for incorrect responses). MR takes two query-response pairs for cross-query comparison on reliability with language models, then aggregates the estimated score of the target query and response. 
    }
    \label{fig:fs-profile-overview}
    % \vspace{-1em}
\end{figure*}

\section{Introduction}

Large language models (LLMs) have demonstrated strong performance in various tasks~\citep{gpt4-tech-report, llama2, du-etal-2022-glm}. However, they still face reliability challenges. For example, these models often produce responses that seem plausible but are factually incorrect, a phenomenon known as ``hallucination''~\citep{hallucination-survey}. Fine-tuning and alignment techniques have been extensively studied to mitigate this issue~\citep{instructgpt, wang-etal-2023-self-instruct, rafailov2023direct, align-for-honesty, gupta2024language}. Recent studies sadly demonstrate that hallucination is inevitable~\citep{hallucination-inevitable}. Consequently, instead of resolving it directly, we focus on developing techniques to discriminate the reliability of responses from LLMs.

Recent research has highlighted the potential of strong LLMs in evaluating response reliability~\citep{zheng2023judging}. Highly capable models, such as GPT-4~\citep{gpt4-tech-report}, have shown effective in assessing the quality of LLM responses through few-shot in-context learning (ICL)~\citep{yin-etal-2023-large}. However, these models are often prohibitively large, resulting in high computational and monetary costs. Also, most of these models are closed-source, which limits their deployment in local environments. On the other hand, weak models, are often better choices for efficiency and local setup. However, their performance is usually lower, probably due to the inherent low capacity in ICL (Figure~\ref{fig:fs-profile-overview} (a)). This raises a critical question: \emph{How can we enable weak LLMs to effectively judge the reliability of LLM responses?}

To address this question, we propose a novel method named \emph{Meta Ranking} (MR). Inspired from the idea of pairwise ranking on responses to the same query~\citep{pandalm, critique-llm, judgelm}, we raised a core hypothesis of MR, that the reliability of a response can be discerned by comparing the query-response pair with other pairs of known reliability. Unlike traditional methods that let an LLM directly judge the response to a query, MR involves cross-query comparison of the target query-response pair with multiple reference pairs (Figure~\ref{fig:fs-profile-overview} (b)). Specifically, MR utilizes a fixed set of query-response pairs with pre-determined scores as reference. For any given target query-response pair, the LLM determines whether this pair is more reliable than each of the reference pairs. A voting mechanism is then employed to aggregate these comparisons and reach a final judgment. Here, ``reliable'' encompasses attributes such as correctness and quality as required by the context. Theoretically, it avoids item perturbation problems in few-shot ICL~\citep{few-shot-permutation} and over-confidence in judgement~\citep{xiong2024can} for language models. Experimental results demonstrate that MR enables weak LLMs to effectively judge the reliability of LLM responses on reasoning tasks, which is previously limited to those strong LLMs. 

Moreover, we showcase the application of MR with a weak LLM in two practical scenarios for validation: (1) enhancing LLM inference through \emph{model cascading} between open- and closed-source LLMs, where queries are routed to the appropriate LLM based on reliability assessments. It demands high efficiency of the judgement process. MR achieves performance comparable to GPT-4-turbo while consuming less than half API tokens. And (2) iteratively filtering training datasets to improve LLMs during \emph{instruction tuning}, which prefers local deployment. MR leads to advancements over existing data selection methods on the Alpaca-52k dataset~\citep{alpaca}, in terms of effectiveness and efficiency. 

In summary, our contributions are threefold:
\begin{enumerate}
    \item We introduce \emph{Meta Ranking} (MR), a novel method for assessing the reliability of LLM responses through cross-query comparison with reference query-response pairs.
    \item We demonstrate that MR enables weak LLMs to judge the reliability of LLM responses, surpassing previous uncertainty estimation and prompting methods even with some strong LLMs, on effectiveness and efficiency.
    \item Additionally, we explore two practical applications of MR, improving strong LLMs in both inference and training, respectively. These results underscore the considerable potential of our proposed method in both efficiency and effectiveness.
\end{enumerate}

\section{Meta Ranking}\label{sec:method}
\begin{figure*}[t]
    \begin{minipage}{0.35\linewidth}
        \centering
    \includegraphics[width=0.98\linewidth]{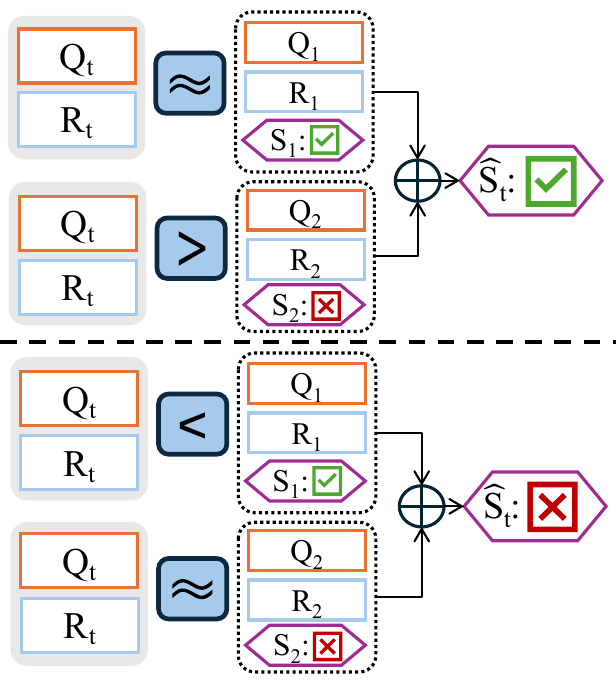}
    \caption{Example illustrations of MR process. The correctness of the target response ($\text{R}_\text{t}$) is considered according to comparisons with reference query-response pairs.}\label{fig:hypothesis}
    \end{minipage}
    \hfill
    \begin{minipage}{0.63\linewidth}
    \centering
    \includegraphics[width=\linewidth]{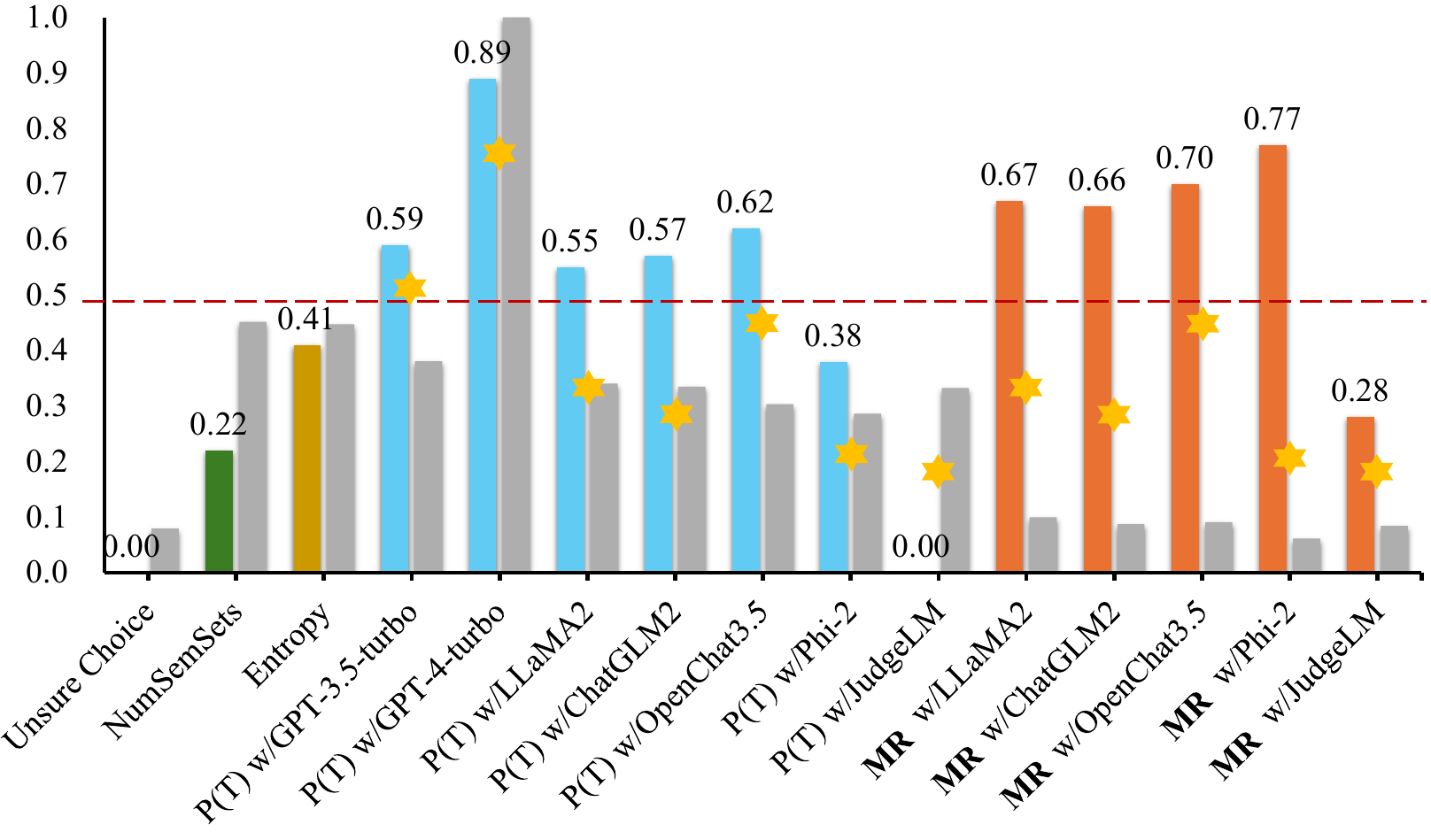}
    \caption{The precision scores and inference time (gray bars) in error detection experiments for target responses from LLaMA-2 on the MMLU dataset. The \textcolor{red}{dashed red line} represents the random selection. We used examples in the development set as reference for few-shot methods. Stars denote the performance on MMLU of underlying LLMs of each method.}\label{fig:error-detect}
    \end{minipage}
    % \vspace{-1em}
\end{figure*}

This section demonstrates how cross-query comparisons could reveal the reliability of the target query-response pair with limited reference examples from the same source LLM. The intuition is as follows: Taking correctness assessment as an example, \emph{the target pair is likely to be correct when ranked closer to a correct reference pair and likely to be incorrect otherwise}, as shown in Figure~\ref{fig:hypothesis}. Below, we outline the specific steps and considerations accordingly. 

Formally, suppose we have $N$ reference query-response pairs
% \vspace{-0.5em}
\begin{equation}
    \mathcal{X}=\left\{(Q_i, R_i, S_i)\right\}, 
\end{equation}
where  $i=1,\cdots,N$, $Q_i$ and $R_i$ are the $i$-th reference query and response, respectively. For each pair of $(Q_i, R_i)$, we have a score $S_i$ that represents its reliability. We aim to evaluate the reliability $S_{\mathrm{t}}$ of a target response $R_{\mathrm{t}}$ to the target query $Q_{\mathrm{t}}$. For binary classification scenarios (e.g., correctness assessment), $S_i, S_{\mathrm{t}}\in\{+1,-1\}$, where $+1$ denotes the response is $\mathrm{True}$ and $-1$ denotes $\mathrm{False}$.

\paragraph{Cross-Query Comparison}
The basic operation of \textit{Meta Ranking} is to compare the target query-response pair with each of the reference query-response pairs. For brevity, we denote the target query-response pair as $P_{\mathrm{t}}=(Q_{\mathrm{t}}, R_{\mathrm{t}})$, and the $i$-th reference query-response pair as $P_i=(Q_i, R_i)$. Then, the cross-query comparison operation and its result are denoted as follows:
% \vspace{-0.5em}
\begin{equation}
    r_i=\mathrm{MR}\left(P_{\mathrm{t}}, P_i\right), i=1,\cdots,N, 
\end{equation}
where $r_i\in\{\pm1, 0\}$, $+1$, $0$, and $-1$ denote the target pair is better than, equal to, or worse than the $i$-th reference pair, respectively. In practice, $\mathrm{MR}(\cdot,\cdot)$ is implemented by directly prompting LLMs or using the relative magnitude of quality estimation scores of each response to its query. 

\paragraph{Aggregation}
The final judgement is obtained by aggregating the comparison results to arrive at the estimated reliability score of the target query-response pair. Specifically, we will upvote if the target air is ranked higher than a reference pair, i.e., $r_i=+1$, and downvote when ranked lower. Also, ranking higher than a correct reference and than an incorrect reference will result in different voting values. For each comparison between the target and the $i$-th reference pair, the individual voting value is 
\begin{equation}\label{equ:si-def}
s_i = S_i\cdot \delta_{\mathrm{sgn}(S_i)\cdot r_i}, i=1,\cdots,N, 
\end{equation}
where $\mathrm{sgn}(S_i)\cdot r_i \in \{\pm1, 0\}$, and thus $\delta_{\pm1}$ and $\delta_{0}$ are hyperparameters. 
For instance, in terms of correctness, $\delta_{+1}$ is the absolute voting value when the target pair is ranked higher than a correct reference ($\mathrm{sgn}(S_i)=r_i=+1$), or lower than an incorrect one ($\mathrm{sgn}(S_i)=r_i=-1$). Note that we require that $\delta_{+1}>0, \delta_{-1}<0$.
Formally, we denote the total vote value as $s$: 
% \vspace{-0.25em}
\begin{equation}\label{equ:mr}
    s=\sum_{i=1}^N s_i = \sum_{i=1}^N S_i\cdot \delta_{\mathrm{sgn}(S_i)\cdot r_i}, 
\end{equation}
And we say the target response is reliable if $s\geq0$ and unreliable otherwise. Thus, the estimated target reliability score $\widehat{S_{\mathrm{t}}}\approx\mathrm{sgn}(s)$ for correctness assessment. The entire algorithmic process is shown in Appendix~\ref{app:b-impl}. In practice, $N$ is usually small due to efficiency and the limited labeled data. 

For theoretical validation, when cross-query comparison reveals the actual relation between $S_i$ and $S_{\mathrm{t}}$, we show that $\mathrm{sgn}(\delta_{r_i})\equiv\mathrm{sgn}(S_{\mathrm{t}}-S_i)$ under reasonable constraints in Appendix~\ref{app:proof-sr}, where $\mathrm{sgn}(\cdot)$ is the sign function. 
Thus, $\mathrm{sgn}(s) \approx \mathrm{sgn}(S_{\mathrm{t}} - S_{\mathrm{avg}})$, where $S_{\mathrm{avg}}=\frac{1}{N}\sum_iS_i$. Hence, a negative $s$ means subpar reliability of the target response, and vice versa. 

Under the formulation, there are several interesting properties of MR. First, MR is model-agnostic and permutation-agnostic towards references, which is different from few-shot ICL methods that fluctuate with the order of examples~\citep{few-shot-permutation}. Second, MR avoids the over-confidence issue in LLM judgement~\citep{xiong2024can}, because each reference pair is considered equally with the target pair. 
Moreover, MR could be extended to continuous metrics ($S_i\in\mathbb{R}$, e.g., BLEU~\citep{papineni-etal-2002-bleu}) directly without modification,
and the final judgement of the reliability is still determined by the sign of $s$.

% \vspace{-0.25em}
\section{Main Experiment: Error Detection with Meta Ranking}\label{sec:error-detect-exp}
% \vspace{-0.25em}

In the following section, we empirically demonstrate that \textit{Meta Ranking} can effectively judge the reliability of LLM responses concerning correctness. We leverage error detection on responses generated by LLMs for reasoning tasks for validation. Our findings indicate that the MR approach maintains consistent efficacy across varying base models, response accuracies, and languages.

\subsection{Settings}

The error detection task requires identifying whether a response from an LLM is incorrect given the query. The settings are as follows, and implementation details are in Appendix~\ref{app:b-impl-1}:

\noindent
\textbf{Datasets}: We randomly selected two subjects of each category in the MMLU dataset~\citep{hendrycks2021measuring, hendrycks2021aligning} and one subject of each category in the CMMLU dataset~\citep{li2023cmmlu} as the Chinese benchmark. The datasets contain multi-choice questions from various areas, and each subject has five examples in the development set. 

\noindent
\textbf{Evaluation Metrics}: We adopted micro scores as the main metric to report, that scores are calculated across the whole MMLU or CMMLU dataset. We report precision for performance and seconds per iteration for inference time on a single A800 GPU (Figure~\ref{fig:error-detect}). Inference time are normalized that P(T) with GPT-4-turbo is the unit. And F1 scores are in Appendix~\ref{app:b-impl-1}. AUROC-style metrics are not applicable because the MR algorithm uses a static threshold 0. 

\noindent
\textbf{Response Generation}: For different accuracy levels, we chose \texttt{LLaMA-2-chat-7B}~\citep{llama2} and \texttt{OpenChat-3.5}~\citep{openchat} to generate responses for English questions, and \texttt{ChatGLM-2-6B}~\citep{zeng2023glmb} and \texttt{Yi-6B-Chat}~\citep{yi-01ai} for Chinese ones. We first performed the target response generation on the test set, and then used the same model to generate responses for reference on the development set.

\noindent
\textbf{MR Settings}: We prompted LLaMA-2, ChatGLM-2, {OpenChat-3.5}, GPT-3.5-turbo, and Phi-2 to judge on different query-response pairs with \textit{Meta Ranking}. We also tested an LLM-as-a-Judge-tuned model \texttt{JudgeLM-7B-v1}~\citep{judgelm} in MR to see if the fine-tuning for the evaluation of responses to the same query helps. 
By setting each label $S_i$ with the value $\pm1$ ($\{\mathrm{True}, \mathrm{False}\} \sim \{+1, -1\}$), we apply MR on this task by identifying incorrect responses depending on MR results on the query-response pairs according to Algorithm~\ref{alg:few-shot-new}. 

\noindent
\textbf{Baselines}: 
We compare our method against several baselines with few-shot ICL to ensure a comprehensive evaluation, including (1) appending an \textbf{Unsure Choice}~\citep{p-true}, (2) a black-box uncertainty estimation method \textbf{NumSemSets}~\citep{gen-confidence}, (3) a white-box method \textbf{Entropy}~\citep{uncertainty-aware}, and (4) P(True) (\textbf{P(T)})~\citep{p-true} which directly asks an LLM about the correctness of a query-response pair.

\begin{figure*}[t]

    \begin{minipage}{0.57\linewidth}
        \centering
        \captionof{table}{The micro precision scores on error detection experiments on the MMLU and CMMLU datasets with responses generated by different LLMs. The \textbf{bold font} denotes best results. LLMs in the second row of the header are sources of responses.}
        \label{tab:error-detect-tasks}
        \resizebox{\linewidth}{!}{
        \begin{tabular}{lcccc}
        \toprule
        \multirow{2}{*}{\textbf{Method}}   &  \multicolumn{2}{c}{\textbf{MMLU} (En)}  &   \multicolumn{2}{c}{\textbf{CMMLU} (Zh)}   \\ \cmidrule(lr){2-3}\cmidrule(lr){4-5}
             &    LLaMA-2   &  OpenChat-3.5  &  ChatGLM-2  &  Yi \\
             \midrule\midrule
             \textcolor{gray}{\textit{Random Selection}}  &  \textcolor{gray}{\textit{0.50}}  &  \textcolor{gray}{\textit{0.50}}  &  \textcolor{gray}{\textit{0.50}}  &  \textcolor{gray}{\textit{0.50}} \\\midrule
         P(T)  w/ OpenChat-3.5  &   0.62   &  0.38   &   0.31  &  0.35  \\
       P(T)  w/ GPT-3.5-turbo  &   0.59   &   0.65   &   0.22  &  0.21  \\\midrule
        \textbf{MR} w/  Phi-2  &   0.77   &   0.73   &  0.69   &  0.52  \\
                 \textbf{MR}   w/ GPT-3.5-turbo  &   \textbf{0.78}   &   \textbf{0.79}   &   \textbf{0.75}   &  \textbf{0.73} \\
             \bottomrule
        \end{tabular}}

    \end{minipage}
    \hfill
    \begin{minipage}{0.42\linewidth}
        \centering
        \includegraphics[width=\linewidth]{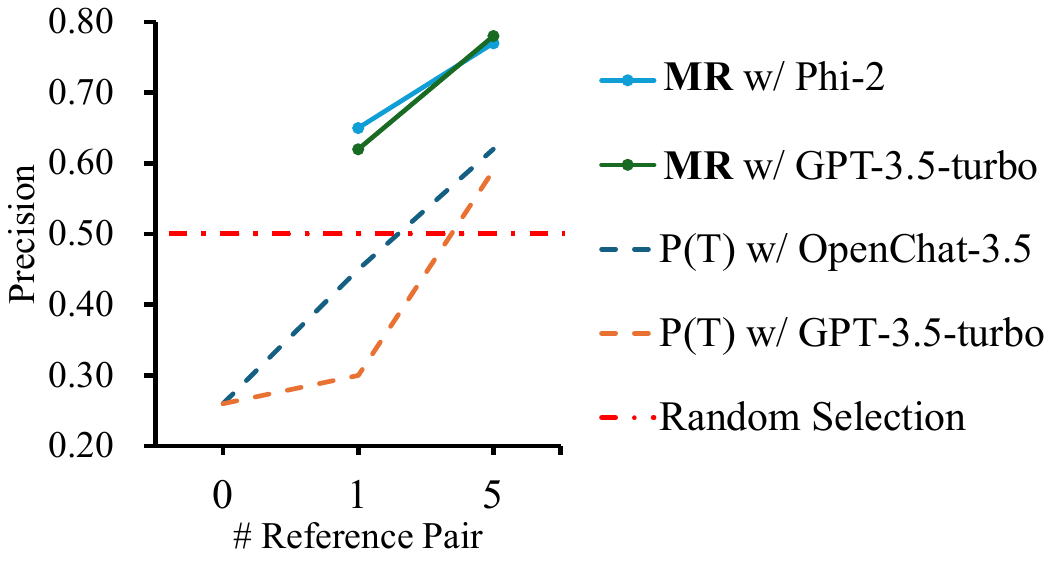}
        \caption{The change of precision scores with the number of reference pairs on the MMLU dataset with target responses from LLaMA-2.}
        \label{fig:fewshot-size}
    \end{minipage}
    % \vspace{-1em}
\end{figure*}

\subsection{Discussion}\label{sec:ed-discuss}

\paragraph{Effectiveness of Meta Ranking} 
MR is effective across different LLM backbones, which might attribute to the position agnosticism to reference and judgement without overconfidence. In Figure~\ref{fig:error-detect}, we report results of all baselines and MR in error detection on LLaMA-2-generated responses, and the actual performance of LLMs on MMLU. Impressively, we found that MR with Phi-2 notably exceeds all baseline methods, except for P(T) with GPT-4-turbo, reaching a precision score of 0.77, more than double the performance of P(T) with Phi-2 and reaching 88\% GPT-4-turbo performance. With LLaMA-2 and ChatGLM-2, MR exceeds P(T) significantly. However, JudgeLM performs not as well as other pretrained or general instruction-tuned LLMs in the MR results and fails to generalize for the P(T) method. It might be the fine-tuning process limits its generalization. We also depict F1 scores in Appendix~\ref{app:b-impl-1} to indicate MR is not biased to identify most responses as incorrect ones. In short, MR consistently outperforms the random baseline to a greater margin than P(T), except for P(T) with GPT-4-turbo. GPT-4-turbo could accurately detect the error, probably relying on its strong reasoning capabilities and generalizability~\citep{gpt4-tech-report}.

\paragraph{Performance of MR with Weak LLMs}
In Figure~\ref{fig:error-detect}, the performance of P(T) across different models displays a positive correlation with their actual performance on reasoning tasks, while MR has demonstrated strong robustness across models with different capabilities. 
We further investigate the performance of LLMs with different capabilities across languages and tasks. 
In Table~\ref{tab:error-detect-tasks}, we select top performing P(T) and MR methods and report the precision score across all target responses on the MMLU and CMMLU datasets. We omit GPT-4-turbo for the cost and its unparalleled capabilities to open-source models we have tested. From results, P(T) with OpenChat-3.5 performs worse than random selection when facing more accurate responses on MMLU. 
In contrast, MR shows significant generalizability with weak models, e.g., the 2.7B Phi-2, across different accuracy levels of responses, impressively surpassing P(T) w/ GPT-3.5-turbo and OpenChat-3.5 on responses in different accuracy levels. 
It positively indicates the effectiveness of weak LLMs in accurately detecting reasoning errors in LLM responses. 

\paragraph{Impact of the Number of Reference Pairs} 
In Figure~\ref{fig:fewshot-size}, we illustrate the change of precision scores with the number of reference pairs, where MR exhibits that it could function with far less labeled data compared to previous methods. Notably, other uncertainty-based methods are incompatible with the 1-shot setting since there are usually no correct examples for calibration. 
Upon the ablation study, we observe that reducing five examples to one leads to a slightly decreased performance of MR, indicating the robustness of the cross-query comparison. Also, the result highlights the effectiveness of MR with limited labeled data compared to P(T). Without reference examples, P(T) faces a great performance drop even with capable models (e.g., GPT-3.5-turbo and OpenChat-3.5), showing inferior performance to random selection. It is also worth noting that uncertainty-based methods like NumSemSets and Entropy usually require hundreds of examples for calibration on distinguishing errors~\citep{uncertainty-aware, mielke-etal-2022-reducing}, explaining the relative low results of these uncertainty-based methods in Figure~\ref{fig:error-detect} when there are only five examples for reference.

\paragraph{Performance on Non-English Tasks}
In Table~\ref{tab:error-detect-tasks}, the overall results on Chinese reasoning problems are significantly lower than on English ones, demonstrating that non-English languages do influence error detection performance. However, MR exhibits strong robustness across languages, though P(T) performs worse than random selection in all CMMLU results. 
Please also see Appendix~\ref{app:low-res} for experiments on Japanese as a representative low-resource language, which causes much lower performance in judgement due to multilingual capacities of LLMs.

\begin{figure*}[t]
    \centering
    \includegraphics[width=0.95\linewidth]{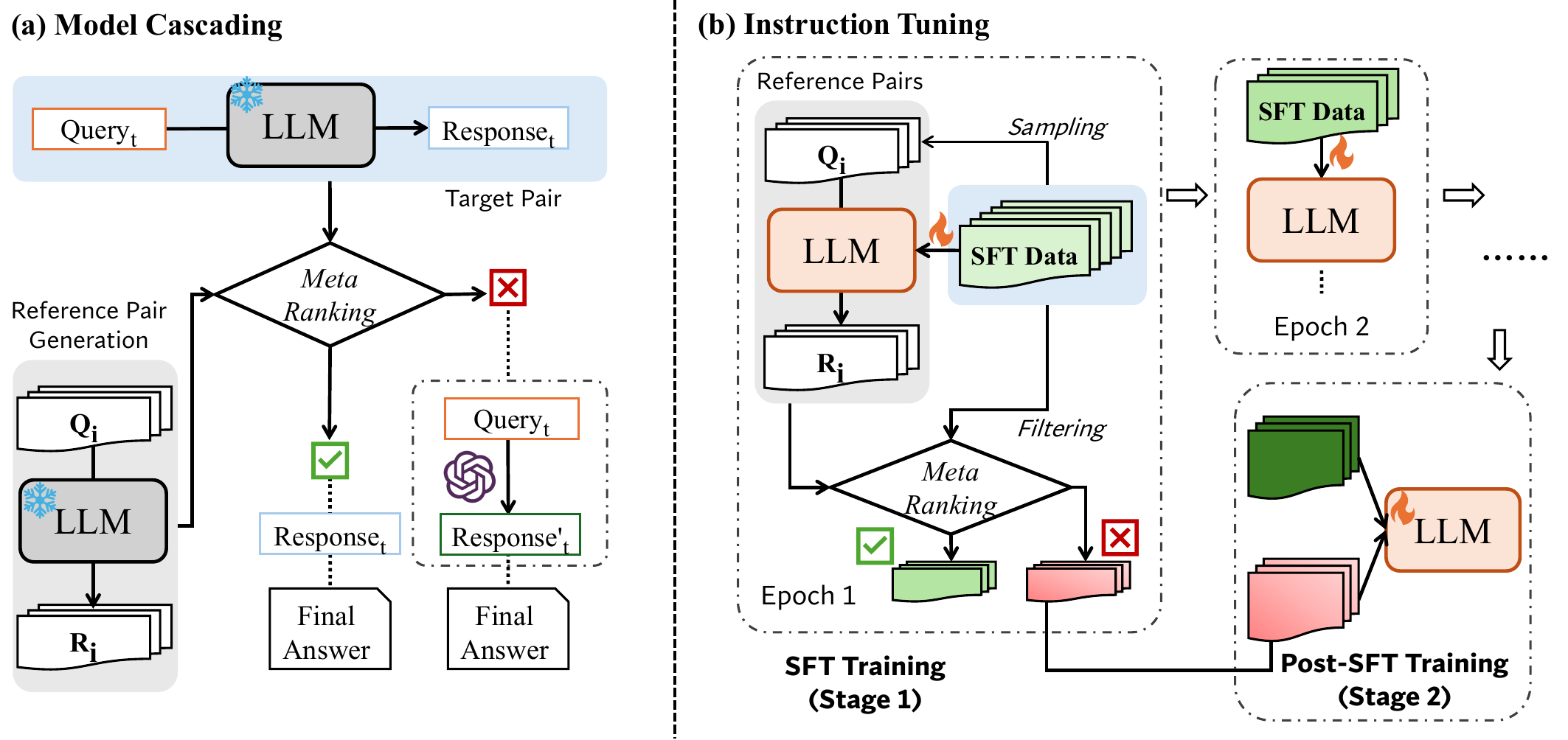}
    \caption{Two applications of \textit{Meta Ranking} for inference- and training-time LLM enhancement, respectively. (a) Model Cascading (left): MR identifies reliability of responses and routes unsolved queries from open-source LLMs towards closed-source LLMs for better results. (b) Instruction Tuning (right): MR filters low-quality data after each epoch in SFT and then tune LLMs with low- and high-quality data. MR results depend on the reference pairs generated from the LLM on samples of the training dataset. $Q_{\mathrm{i}}, R_{\mathrm{i}}$ denote reference query-response pairs for the MR algorithm, and Query$_\mathrm{t}$, Response$_\mathrm{t}$ denote the target pair.}\label{fig:apps}
    \vspace{-1em}
\end{figure*}

\section{Applications of Meta Ranking}\label{sec:applications}

In this section, we present two practical applications to further validate the effectiveness of \textit{Meta Ranking}, as shown in Figure~\ref{fig:apps}. Each application is implemented by collecting \textbf{reference query-response pairs} and setting \textbf{reference reliability scores}. (a) With the assessment of the correctness of responses from open-source LLMs to the given queries, we identify and route unsolved queries to stronger closed-source LLMs. It could achieve better efficiency and remaining performance of closed-source models. However, it demands the model used for judgement is weaker than the open-source LLM in deployment, otherwise it is better in place to respond to queries. 
(b) By evaluating the quality of instruction data, we can refine the supervised fine-tuning (SFT) for LLM instruction tuning, whose key factor is the quality of training data~\citep{deita}. By filtering low-quality data after each epoch and further introducing post-SFT training with mere instruction data, we achieve significant improvement against state-of-the-art SFT data selection methods. For data related application, the method is better to be locally deployed. 
Thus, we decide MR with weak LLMs is suitable for judgement for these. 

\subsection{Model Cascading}\label{sec:query-routing}

Since LLMs exhibit varying degrees of accuracy across various tasks, we propose using MR within a model cascading system. As depicted in Figure~\ref{fig:apps} (a), this system employs MR to assess the reliability of generated responses from open-source LLMs. Queries with responses deemed unreliable by MR are routed to more powerful, but also more costly, closed-source LLMs for better answers. This system aims to achieve performance similar to that of closed-source models with improved efficiency.

\subsubsection{Implementation}

Assuming both the development and the test sets are drawn from the same underlying distribution. Given that the MR method requires reference query-response pairs, we first feed the queries in the development set to the open-source LLM and evaluate the generated responses against the ground truth. 
Formally, for every query $Q_i$, let $R_i^{(\theta)}$ represent the response generated by an open-source LLM (parameterized by $\theta$), and $S_i^{(\theta)}$ denote the evaluation result according to the ground truth to $Q_i$ and an appropriate metric. By applying the model to each query in the development set with $N$ samples, we produce responses $\left\{R_i^{(\theta)}\right\}_{i=1}^{N}$ and form a set of \textbf{reference query-response pairs} $\mathcal{X} = \left\{P_i^{(\theta)} = (Q_i, R_i^{(\theta)})\right\}_{i=1}^{N}$, along with associated evaluation results $\left\{S_i^{(\theta)}\right\}_{i=1}^{N}$.

There are \textbf{two reasonable ways to derive reliability scores} for $\mathcal{X}$ in MR. 
The first is to directly define $S_i \triangleq S_i^{(\theta)}$, which we term MR$(\theta)$. 
The second option is to compute the responses from the closed-source LLM ($\Theta$) and their evaluation results $\left\{S_i^{(\Theta)}\right\}_{i=1}^{N}$. We could define the label as follows:
\begin{equation}\label{equ:ci}
    S_i \triangleq S_i^{(\theta)} - S_i^{(\Theta)}, 
\end{equation}
which grants positive scores only when the response from open-source LLMs is better than the one from closed-source LLM, since the model cascading causes performance improvement vice versa. We denote it by MR$(\Delta)$, which considers the gap between open- and closed-source LLMs. 

Accordingly, we can obtain estimated reliability for $P_t^{(\theta)}$ from the MR approach during inference on test sets. If the assessment indicates $R_t$ as an unreliable response, we direct $Q_t$ to a closed-source large language model (such as GPT-4-turbo) to secure a more precise response. Conversely, we preserve the original response for a positive MR result. Ideally, this approach can enhance performance with moderate costs, as it is generally observed that a poorly accurate response from open-source LLMs often corresponds to a difficult query, which requires strong LLMs to respond, and vice versa.

\begin{table*}[t]
    \small
    \centering
    \caption{The overall results in model cascading experiments. All routing strategies are implemented with Phi-2, whose results are \textcolor{gray}{gray}. ``Overall'' denotes the macro average value across tasks. The \textbf{bold font} denotes the best result using model cascading and the \uline{underlined numbers} denote the best result for each setting. The number in the parentheses denotes the improvement over the best among open-source LLMs and the ensemble baseline without model cascading. For notation, we use serial numbers to represent LLMs, e.g., ``\ding{172}/\ding{173} $+$ \ding{174}'' represents model cascading from LLaMA-2 or ChatGLM-2 to GPT-3.5-turbo. }
    \label{tab:query-routing}
    \resizebox{\linewidth}{!}{
    \begin{tabular}{llcccclc}
    \toprule
    \multirow{2}{*}{\textbf{Model}} & \textbf{Routing}  & \multicolumn{2}{c}{\textbf{Reasoning}} & \multicolumn{2}{c}{\textbf{Translation}} & \multirow{2}{*}{\textbf{Average}} &\textbf{\#Token (API)}  \\ \cmidrule(lr){3-4}\cmidrule(lr){5-6}
      & \textbf{Strategy}  & \textbf{English}  &  \textbf{Chinese}  &  \textbf{Zh-En}  &  \textbf{En-Zh}  &   &  (Relative Value) \\  
    \midrule\midrule
    \textcolor{gray}{\textit{Phi-2}}  & \textcolor{gray}{-}  & \textcolor{gray}{\textit{22.48}} & \textcolor{gray}{\textit{24.84}} & \textcolor{gray}{\textit{42.32}} & \textcolor{gray}{\textit{23.18}}  &  \textcolor{gray}{\textit{28.21}}  & \textcolor{gray}{-} \\ \midrule\midrule
    LLaMA-2 (\ding{172}) & \multirow{4}{*}{-}  &  34.37  &  32.86 &  58.53  &  51.33  & 44.27 & \multirow{3}{*}{-} \\
     ChatGLM-2  (\ding{173}) & & 30.48   &  49.71 &  58.43 & 63.30 & 50.48 &  \\
    \texttt{Ensemble} (\ding{172}\&\ding{173}) &  & 35.44 &  43.47  & 31.35  & 46.54 &  39.20  & \\ 
     GPT-3.5-turbo (\ding{174}) &  &   \uline{52.91}  &  \uline{54.09}  & \uline{63.67} & \uline{69.14} & \uline{59.95}  &  1.00 \\
    \midrule
    \multirow{4}{*}{\ding{172}/\ding{173} $+$ \ding{174}}&  \texttt{Entropy}  &  38.75  &   47.03  &  58.32   &  65.45   &  52.39 (\textcolor{mydarkgreen}{+1.91}) &  0.38  \\
    &  \texttt{Random}   &  41.89  &  48.32  &  58.97  &  64.83  &  53.50 (\textcolor{mydarkgreen}{+3.02}) &  0.45 \\
    &   \textbf{MR}$(\Delta)$    &  44.30  &  48.67  &  59.80  &   65.87 &  54.66 (\textcolor{mydarkgreen}{+4.18})  &  \textbf{0.24} \\
    &     \textbf{MR}$(\theta)$  &  \textbf{48.62} &  \textbf{52.76}  & \textbf{61.36}   &  \textbf{67.10} & \textbf{57.46} (\textcolor{mydarkgreen}{+\textbf{6.98}})    &  0.42 \\
    \midrule\midrule
    OpenChat-3.5 (\ding{175}) & \multirow{4}{*}{-}  &  45.42  & 40.19 & 61.35 & 60.77  & 51.93 & \multirow{3}{*}{-}  \\
     Yi  (\ding{176}) &   &  42.08  &  61.96 &  60.87  &  62.07 & 56.74  & \\
    \texttt{Ensemble} (\ding{175}\&\ding{176}) &   &  45.68   &  48.24  &  11.16  &  62.90  & 41.99 & \\ 
     GPT-4-turbo (\ding{177}) &   &  \uline{72.86}  &   \uline{62.82}  & \uline{64.73} & \uline{69.95} & \uline{67.59} &  1.00  \\
    \midrule
    \multirow{3}{*}{\ding{175}/\ding{176} $+$ \ding{177}}&  \texttt{Random}  &  46.64  &  \textbf{61.96}  &  61.29  & 63.74 &  58.41 (\textcolor{mydarkgreen}{+1.67}) & 0.44 \\
    &   \textbf{MR}$(\Delta)$  &  57.68  &  61.93 &  61.60  &  67.61 & 62.21 (\textcolor{mydarkgreen}{+5.47})  & \textbf{0.23} \\
    &     \textbf{MR}$(\theta)$ &  \textbf{64.68}  &  61.93 & \textbf{62.60}  &  \textbf{68.11}  & \textbf{64.33} (\textcolor{mydarkgreen}{+\textbf{7.59}})  & 0.43 \\
    \bottomrule
    \end{tabular}
    }
    % \vspace{-1em}
\end{table*}

\subsubsection{Experiment}

\paragraph{Settings}
We leverage reasoning and translation tasks to validate the effectiveness of model cascading. We use the same datasets for reasoning tasks as Section~\ref{sec:error-detect-exp} and randomly sampled test and development set from the FLORES-200 dataset~\citep{nllb-22}. For MR, we prompt Phi-2 in reasoning tasks, and in translation tasks, we adopt \texttt{wmt22-cometkiwi-da}~\citep{rei-etal-2022-cometkiwi} for reference-free quality estimation and thus compare the estimated scores between translations. Due to the diverse LLM capabilities and language biases, we have tested two combinations: (1) LLaMA-2 for English and ChatGLM-2 for Chinese tasks, with GPT-3.5-turbo as the closed-source model; (2) OpenChat-3.5 for English and Yi for Chinese problems, with GPT-4-turbo as the API. 
For baselines, we validate the logits ensemble of the open-source models and implement model cascading with strategies of entropy-based uncertainty estimation~\citep{uncertainty-aware}. We omit the result of the latter for OpenChat-3.5 and Yi since it results in almost no routed queries, probably because the uncertainty threshold is too high to determine false answers. 

\paragraph{Effectiveness of Cascading Guidance from MR}
In Table~\ref{tab:query-routing}, we report the average accuracy on reasoning tasks and the average among BLEU, BLEURT~\citep{sellam-etal-2020-bleurt, pu-etal-2021-learning}, and COMET~\citep{rei-etal-2022-comet} scores on translation tasks. With the model cascading approach, we observe significant improvement against single open-source LLMs. MR$(\theta)$ and MR$(\Delta)$ manage to gain the highest performance improvement across tasks and languages with less than half token consumption. 
Moreover, MR$(\Delta)$ consistently outperforms random selection with nearly half of the token consumption, demonstrating the effectiveness of Equation~\eqref{equ:ci}. We also demonstrate actual deployment costs and inference speeds in Appendix~\ref{app:qr-detail} and found MR-based model cascading costs much lower in monetary expenditure.

\paragraph{Relation with Error Detection Performance}
The performance of our model cascading mechanism is closely related to the effectiveness of error detection. For instance, in Table~\ref{tab:query-routing}, MR outperforms Random and Entropy on absolute performance and token consumption, which denotes MR detects errors in a larger quantity and with higher precision, respectively.  
It suggests that error detection with MR is also robust on open-ended generation tasks with continuous labels, e.g., translation.

\subsection{Instruction Tuning}\label{sec:refine-train}

Recent studies show that the quality of instruction data is essential to SFT performance~\citep{zhou2023lima}. For better instruction tuning on LLMs, we introduce an iterative training data filtering process based on \emph{Meta Ranking} and a post-SFT training stage, as shown in Figure~\ref{fig:apps} (b). The basic intuition is to continuously filter low-quality data, letting LLMs concisely learn from more reliable and fewer training samples at the first stage, and utilize less reliable data samples at the second stage for contrastive learning. MR makes it possible by judging the quality of instruction data rapidly with generated responses from an LLM that reflect its capabilities during training, and is better for local deployment with weak LLMs.

\subsubsection{Implementation}\label{sec:data-refine-impl}

The application contains two stages (Figure~\ref{fig:apps} (b)): (1) SFT with MR guided data selection and (2) post-SFT training with both estimated low- and high-quality data from the last epoch at stage 1.  

For the first stage, besides regular SFT, we extract a small set of queries from the training set and, after each epoch, ask the tuned LLM to respond to those queries. With \textbf{the generated responses and the queries as reference}, the MR method could judge whether a single sample in the original training dataset matches the quality of the reference. For simplicity in MR, we set \textbf{the reliability score of all reference pairs} to 1. Thus, we could filter training data samples that fail the judgment, i.e. unreliable, improving training efficiency and, potentially, LLM performance. 

For the second stage, we want to utilize both the filtered low-quality and the high-quality data to further train the LLM. 
Since post-SFT training methods (e.g., PPO~\citep{instructgpt}, DPO~\citep{rafailov2023direct}) require multiple responses of diverse human preferences or quality to the same query, they are not compatible. Recently, \citet{kto} proposed Kahneman-Tversky Optimization (KTO) to align LLMs towards desired and away from undesired query-response pairs contrastively.  
However, we are aware that their objective is misaligned with our requirement because the low-quality data is derived from the SFT dataset, which is not completely negative. Therefore, in order to incorporate both high- and low-quality data as positive and partially positive samples, we propose positive-KTO (pKTO). Intuitively, pKTO differs from KTO only in dealing with low-quality data, where pKTO regulates the reward of these data with MSE loss instead of decreasing it unlimitedly. 
% which might prevent LLMs from unlearning the inherent knowledge
Please refer to Appendix~\ref{app:pkto-comp} for the detailed implementation and comparisons with DPO and KTO.

\begin{figure*}[t]
    \begin{minipage}{0.56\linewidth}
    \small
    \centering
    \captionof{table}{Results on instruction tuning experiments, where MR is implemented with Phi-2. The \textbf{bold font} denotes best results. ``Full'' denotes the whole dataset.}
    \label{tab:ref-training}
    \resizebox{\linewidth}{!}{
    \begin{tabular}{lccc}
    \toprule
    \textbf{Method} & \textbf{MT-Bench}  & \textbf{AlpacaEval 2.0} & \textbf{\#Token} (M) \\
    \midrule\midrule
     \texttt{Alpaca-13B} & 4.53 & 2.65 & -\\
     \texttt{Guanaco-13B} & - &  3.47  & - \\
     \texttt{Phi-2} & 4.52  & 2.34  & - \\ 
      \texttt{LLaMA-7B} &  2.62  &  0.43 & - \\
    \midrule
    \multicolumn{4}{c}{\textit{Phi-2-Based Results}} \\
    \midrule
     Full & 4.42 & 3.26 & 13.293 \\
     Longest  &  4.56  &  3.32 & \phantom{(}1.008  \\
     Deita  & 4.33 & 3.18  & \phantom{(}9.609\\
     Deita (9k) &  4.64  &   3.29 & \phantom{(}3.981 \\
     \textbf{MR} (Stage 1) & 4.70 & \textbf{3.56}  & \phantom{(}7.509 \\ 
      + \textbf{pKTO} (Stage 2) & \textbf{4.77} & 3.47  & \phantom{(}1.205 \\ 
    \midrule
    \multicolumn{4}{c}{\textit{LLaMA-7B-Based Results}} \\
    \midrule
     Full &  4.36  & 2.53  & 13.293 \\
     Longest  &  4.18  &  2.35 & \phantom{(}1.008 \\
     Deita  & 4.37 & 2.60  & \phantom{(}9.609  \\
     Deita (9k) &  4.48  &  2.86 & \phantom{(}3.981 \\
     \textbf{MR} (Stage 1) & 4.52 & 2.93  & \phantom{(}2.412 \\
      + \textbf{pKTO} (Stage 2) &  \textbf{4.69}  &  \textbf{3.24} & \phantom{(}0.907 \\
    \bottomrule
    \end{tabular}
    }
    \end{minipage}
    \hfill
    \begin{minipage}{0.42\linewidth}
    \small
    \centering
    \captionof{table}{The multi-turn evaluation results from MT-Bench on instruction tuning experiments, which is unfolded from the second column in Table~\ref{tab:ref-training}. ``Avg.'' denotes the average score of different turns in MT-Bench. The \textbf{bold font} denotes the best result for each base model.}
    \label{tab:multi-turn}
    \resizebox{\linewidth}{!}{
    \begin{tabular}{lccc}
    \toprule
    \multirow{2}{*}{\textbf{Method}} & \multicolumn{3}{c}{\textbf{MT-Bench}} \\ \cmidrule(lr){2-4}
     & Turn 1  & Turn 2  &  Avg. \\
    \midrule\midrule
    \texttt{LLaMA-13B} & 3.26 & 1.95 & 2.61 \\
    \texttt{Phi-2} & 6.37 & 2.66 & 4.52 \\
     \texttt{Alpaca-13B} & 4.98 &  4.09  &  4.53 \\
     \texttt{Falcon-40B-instruct} & 5.81 & 4.53  &  5.17 \\

    \midrule
    \multicolumn{4}{c}{\textit{Phi-2-Based post-SFT Results}} \\
    \midrule
    \textbf{MR} (Stage 1)  &   \textbf{6.33}  &  3.08  &  4.70 \\
     $\quad$ + KTO \phantom{((} (Stage 2) & 6.12 &  2.94  &  4.53   \\
     $\quad$ +  \textbf{pKTO} (Stage 2) & 6.27 &  \textbf{3.26} &  \textbf{4.77} \\
    \midrule
    \multicolumn{4}{c}{\textit{LLaMA-7B-Based post-SFT Results}} \\
    \midrule
     \textbf{MR}  (Stage 1) &  \textbf{5.47}  &  3.58  &  4.52\\
     $\quad$ + KTO \phantom{((} (Stage 2)   &  5.15  &  3.60 &  4.38 \\
    $\quad$ + \textbf{pKTO} (Stage 2) & 5.45 & \textbf{3.93}  &  \textbf{4.69} \\
    \bottomrule
    \end{tabular}
    }
    \end{minipage}
    % \vspace{-0.6em}
\end{figure*}

\subsubsection{Experiment}

\paragraph{Settings}
In all experiments, we only use the Alpaca-52k~\citep{alpaca} dataset with 52,002 samples from \texttt{text-davinci-003}~\citep{gpt3}, which is also the target pairs for MR. We utilize AlpacaEval 2.0~\citep{alpaca_eval} and MT-Bench~\citep{zheng2023judging} to benchmark instruction following capabilities. We select strong baselines on SFT data selection, including Deita~\citep{deita} and Longest \citep{long-is-more}. Please refer to Appendix~\ref{app:b-impl-3} for implementation details. For base models, we choose \texttt{Phi-2} and \texttt{LLaMA-7B}~\citep{llama} for instruction tuning. For Phi-2, our method starts with the original 52k dataset. For LLaMA, \citet{chen2024alpagasus} empirically find a 9k subset of Alpaca is the most proper for SFT. We thus adopt the scorer from Deita and extract the top 9k data, noted by Deita (9k), as the initial training set for LLaMA at stage 1. 

\paragraph{High-Quality SFT Data Filtering Guided by MR (Stage 1)}
We report overall results and training tokens (calculated by LLaMA tokenizer) in Table~\ref{tab:ref-training}. For each base model, we found MR guided iterative training data filtering leads to significant improvement in both benchmarks with fewer training tokens. In the first stage of MR guided data selection, our method iteratively filters low-quality data after each epoch. After stage 1, Phi-2 and LLaMA already surpass all baselines, indicating MR effectively picks high-quality training samples in a curriculum way~\citep{curriculum}, which helps align LLMs better compared to selecting data at the beginning for SFT in baseline methods. We also observed significant lower data processing time of MR compared to Deita in Appendix~\ref{app:b-impl-3}. 

\paragraph{Post-SFT Training through MR-Filtered Data (Stage 2)}
We also notice a significant enhancement of stage 2 in Table~\ref{tab:ref-training}. 
We report the detailed scores on each turn of LLM responses from MT-Bench in Table~\ref{tab:multi-turn}. From empirical results, pKTO enhances the second-turn communication of LLMs to a great extent while preserving the instruction-following abilities from SFT at stage 1, which is indicated by the slight drops in first-turn scores. In contrast, KTO fails at this setting, which is aligned with our hypothesis in Section~\ref{sec:data-refine-impl}. By incorporating the low- and high-quality data distinguished from MR, stage 2 further elicits LLMs' capacity, especially for multi-turn scenarios. 

\section{Related Work}\label{sec:related-work}

\paragraph{Evaluation of LLM Responses}
Extensive research has been conducted to evaluate responses from LLMs. Studies have focused on estimating uncertainty to gauge the potential reliability of LLM responses~\citep{kuhn2023semantic, rafailov2023direct}. Furthermore, LLMs are capable of providing uncertainty scores from itself by fine-tuning~\citep{chen-etal-2023-adaptation, gupta2024language}, which usually requires an amount of training data, or black-box measurements~\citep{gen-confidence}. However, these methods often require an amount of labeled data for calibration to determine a threshold~\citep{uncertainty-aware}. Additionally, the LLM-as-a-judge approach effectively assesses the accuracy of LLM responses from strong LLMs~\citep{zheng2023judging} with manual prompting rules. Contrarily, our \textit{Meta Ranking} method leverages weak LLMs and a training-free judgment based on cross-query comparisons with much fewer examples.

\paragraph{Model Cascading with LLMs}
Recent studies on model cascading focus on how LLMs can selectively call tools or stronger models only in difficult situations for better efficiency. Tool calls or another trial happens on external feedback from environment~\citep{lin2023swiftsage, shinn2023reflexion}. For tasks with explicit criteria, e.g., coding, LLMs can call stronger models after their failure~\citep{zhang2023ecoassistant, yue2024large}. Selection can also be achieved through fine-tuning~\citep{uncertainty-api} or uncertainty estimation~\citep{uncertainty-aware, gupta2024language}. In our approach, MR decides to route queries on complicated open-ended tasks, and we also validate it on coding tasks in Appendix~\ref{app:code-exp}.

\paragraph{Data-Efficient Training for LLMs}
Coresets~\citep{pmlr-v119-mirzasoleiman20a} are used in machine learning for a long period. For LLMs, several data selection methods are developed for SFT~\citep{deita, zhou2023lima, nuggets, chen2024alpagasus} and post-SFT stages~\citep{rest, rest-meet-react}. Inspired by the latter, we introduce MR guided instruction tuning, where training data are filtered iteratively based on MR results after each epoch, and used in post-SFT training.

\section{Conclusion}

We present \emph{Meta Ranking} (MR), a novel method effectively enabling weak LLMs to judge the reliability of LLM responses. By comparing a target query-response pair with a small number of reference pairs, MR outperforms strong baselines in error detection without fine-tuning. Furthermore, the method significantly enhances strong LLMs' performance and efficiency in two practical application scenarios, model cascading and instruction tuning. These findings highlight the potential of MR for broader inference- and training-time applications with LLMs.

\section*{Limitations}

There are several limitations to our work that we would like to acknowledge:

First, we have not explored deep in the compatibility between the model training process and the \textit{Meta Ranking} (MR) method. Different model training strategies may affect the effectiveness of MR. It is an interesting direction for future work to study how to better integrate MR with the alignment process (e.g., SFT and post-SFT training) on LLMs.

Also, we have not focused on finding potential applications of the \textit{Meta Ranking} method for strong models. Our current experiments focus on enabling weak LLMs to judge the reliability of LLM responses due to its superior efficiency and effectiveness. It remains an open question of what practical usage could incorporate MR with strong models like GPT-4-turbo. Exploring the potential applications of MR for strong models is also a direction for future work.

In conclusion, while our proposed \textit{Meta Ranking} method has shown promising results in enabling weak LLMs to judge the reliability of individual responses and enhancing LLM performance in practical applications, there are still spaces to be explored. We hope that future research can address these limitations and further improve the method. 

\bibliography{anthology,custom}

\begin{thebibliography}{63}
\expandafter\ifx\csname natexlab\endcsname\relax\def\natexlab#1{#1}\fi

\bibitem[{01.AI(2023)}]{yi-01ai}
01.AI. 2023.
\newblock {Yi: Building the Next Generation of Open-Source and Bilingual LLMs}.
\newblock \url{https://github.com/01-ai/Yi}.

\bibitem[{{Aksitov} et~al.(2023){Aksitov}, {Miryoosefi}, {Li}, {Li}, {Babayan}, {Kopparapu}, {Fisher}, {Guo}, {Prakash}, {Srinivasan}, {Zaheer}, {Yu}, and {Kumar}}]{rest-meet-react}
Renat {Aksitov}, Sobhan {Miryoosefi}, Zonglin {Li}, Daliang {Li}, Sheila {Babayan}, Kavya {Kopparapu}, Zachary {Fisher}, Ruiqi {Guo}, Sushant {Prakash}, Pranesh {Srinivasan}, Manzil {Zaheer}, Felix {Yu}, and Sanjiv {Kumar}. 2023.
\newblock \href {https://arxiv.org/abs/2312.10003} {{ReST meets ReAct: Self-Improvement for Multi-Step Reasoning LLM Agent}}.
\newblock \emph{Computing Research Repository}, arXiv:2312.10003.

\bibitem[{Bengio et~al.(2009)Bengio, Louradour, Collobert, and Weston}]{curriculum}
Yoshua Bengio, J\'{e}r\^{o}me Louradour, Ronan Collobert, and Jason Weston. 2009.
\newblock \href {https://doi.org/10.1145/1553374.1553380} {{Curriculum learning}}.
\newblock In \emph{Proceedings of the 26th Annual International Conference on Machine Learning}, ICML '09, page 41–48, New York, NY, USA. Association for Computing Machinery.

\bibitem[{Brown et~al.(2020)Brown, Mann, Ryder, Subbiah, Kaplan, Dhariwal, Neelakantan, Shyam, Sastry, Askell, Agarwal, Herbert-Voss, Krueger, Henighan, Child, Ramesh, Ziegler, Wu, Winter, Hesse, Chen, Sigler, Litwin, Gray, Chess, Clark, Berner, McCandlish, Radford, Sutskever, and Amodei}]{gpt3}
Tom~B. Brown, Benjamin Mann, Nick Ryder, Melanie Subbiah, Jared Kaplan, Prafulla Dhariwal, Arvind Neelakantan, Pranav Shyam, Girish Sastry, Amanda Askell, Sandhini Agarwal, Ariel Herbert-Voss, Gretchen Krueger, Tom Henighan, Rewon Child, Aditya Ramesh, Daniel~M. Ziegler, Jeffrey Wu, Clemens Winter, Christopher Hesse, Mark Chen, Eric Sigler, Mateusz Litwin, Scott Gray, Benjamin Chess, Jack Clark, Christopher Berner, Sam McCandlish, Alec Radford, Ilya Sutskever, and Dario Amodei. 2020.
\newblock {Language models are few-shot learners}.
\newblock In \emph{Proceedings of the 34th International Conference on Neural Information Processing Systems}, NIPS'20, Red Hook, NY, USA. Curran Associates Inc.

\bibitem[{Chen et~al.(2023)Chen, Yoon, Ebrahimi, Arik, Pfister, and Jha}]{chen-etal-2023-adaptation}
Jiefeng Chen, Jinsung Yoon, Sayna Ebrahimi, Sercan Arik, Tomas Pfister, and Somesh Jha. 2023.
\newblock \href {https://doi.org/10.18653/v1/2023.findings-emnlp.345} {Adaptation with self-evaluation to improve selective prediction in {LLM}s}.
\newblock In \emph{Findings of the Association for Computational Linguistics: EMNLP 2023}, pages 5190--5213, Singapore. Association for Computational Linguistics.

\bibitem[{Chen et~al.(2024)Chen, Li, Yan, Wang, Gunaratna, Yadav, Tang, Srinivasan, Zhou, Huang, and Jin}]{chen2024alpagasus}
Lichang Chen, Shiyang Li, Jun Yan, Hai Wang, Kalpa Gunaratna, Vikas Yadav, Zheng Tang, Vijay Srinivasan, Tianyi Zhou, Heng Huang, and Hongxia Jin. 2024.
\newblock \href {https://openreview.net/forum?id=FdVXgSJhvz} {{Alpagasus: Training a Better Alpaca Model with Fewer Data}}.
\newblock In \emph{The Twelfth International Conference on Learning Representations}.

\bibitem[{Du et~al.(2022)Du, Qian, Liu, Ding, Qiu, Yang, and Tang}]{du-etal-2022-glm}
Zhengxiao Du, Yujie Qian, Xiao Liu, Ming Ding, Jiezhong Qiu, Zhilin Yang, and Jie Tang. 2022.
\newblock \href {https://doi.org/10.18653/v1/2022.acl-long.26} {{GLM}: General language model pretraining with autoregressive blank infilling}.
\newblock In \emph{Proceedings of the 60th Annual Meeting of the Association for Computational Linguistics (Volume 1: Long Papers)}, pages 320--335, Dublin, Ireland. Association for Computational Linguistics.

\bibitem[{{Erbacher} et~al.(2024){Erbacher}, {Falissar}, {Guigue}, and {Soulier}}]{uncertainty-api}
Pierre {Erbacher}, Louis {Falissar}, Vincent {Guigue}, and Laure {Soulier}. 2024.
\newblock \href {https://arxiv.org/abs/2401.01780} {{Navigating Uncertainty: Optimizing API Dependency for Hallucination Reduction in Closed-Book Question Answering}}.
\newblock \emph{Computing Research Repository}, arXiv:2401.01780.

\bibitem[{{Ethayarajh} et~al.(2024){Ethayarajh}, {Xu}, {Muennighoff}, {Jurafsky}, and {Kiela}}]{kto}
Kawin {Ethayarajh}, Winnie {Xu}, Niklas {Muennighoff}, Dan {Jurafsky}, and Douwe {Kiela}. 2024.
\newblock \href {https://arxiv.org/abs/2402.01306} {{KTO: Model Alignment as Prospect Theoretic Optimization}}.
\newblock \emph{Computing Research Repository}, arXiv:2402.01306.

\bibitem[{Fadeeva et~al.(2023)Fadeeva, Vashurin, Tsvigun, Vazhentsev, Petrakov, Fedyanin, Vasilev, Goncharova, Panchenko, Panov, Baldwin, and Shelmanov}]{fadeeva-etal-2023-lm}
Ekaterina Fadeeva, Roman Vashurin, Akim Tsvigun, Artem Vazhentsev, Sergey Petrakov, Kirill Fedyanin, Daniil Vasilev, Elizaveta Goncharova, Alexander Panchenko, Maxim Panov, Timothy Baldwin, and Artem Shelmanov. 2023.
\newblock \href {https://doi.org/10.18653/v1/2023.emnlp-demo.41} {{LM}-polygraph: Uncertainty estimation for language models}.
\newblock In \emph{Proceedings of the 2023 Conference on Empirical Methods in Natural Language Processing: System Demonstrations}, pages 446--461, Singapore. Association for Computational Linguistics.

\bibitem[{{Gulcehre} et~al.(2023){Gulcehre}, {Le Paine}, {Srinivasan}, {Konyushkova}, {Weerts}, {Sharma}, {Siddhant}, {Ahern}, {Wang}, {Gu}, {Macherey}, {Doucet}, {Firat}, and {de Freitas}}]{rest}
Caglar {Gulcehre}, Tom {Le Paine}, Srivatsan {Srinivasan}, Ksenia {Konyushkova}, Lotte {Weerts}, Abhishek {Sharma}, Aditya {Siddhant}, Alex {Ahern}, Miaosen {Wang}, Chenjie {Gu}, Wolfgang {Macherey}, Arnaud {Doucet}, Orhan {Firat}, and Nando {de Freitas}. 2023.
\newblock \href {https://arxiv.org/abs/2308.08998} {{Reinforced Self-Training (ReST) for Language Modeling}}.
\newblock \emph{Computing Research Repository}, arXiv:2308.08998.
\newblock Version 2.

\bibitem[{Gupta et~al.(2024)Gupta, Narasimhan, Jitkrittum, Rawat, Menon, and Kumar}]{gupta2024language}
Neha Gupta, Harikrishna Narasimhan, Wittawat Jitkrittum, Ankit~Singh Rawat, Aditya~Krishna Menon, and Sanjiv Kumar. 2024.
\newblock \href {https://openreview.net/forum?id=KgaBScZ4VI} {Language model cascades: Token-level uncertainty and beyond}.
\newblock In \emph{The Twelfth International Conference on Learning Representations}.

\bibitem[{{Han} et~al.(2024){Han}, {Buntine}, and {Shareghi}}]{uncertainty-aware}
Jiuzhou {Han}, Wray {Buntine}, and Ehsan {Shareghi}. 2024.
\newblock \href {https://arxiv.org/abs/2401.14016} {{Towards Uncertainty-Aware Language Agent}}.
\newblock \emph{Computing Research Repository}, arXiv:2401.14016.
\newblock Version 2.

\bibitem[{Hao et~al.(2023)Hao, Liu, Wang, and Hu}]{hao2023toolkengpt}
Shibo Hao, Tianyang Liu, Zhen Wang, and Zhiting Hu. 2023.
\newblock \href {https://openreview.net/forum?id=BHXsb69bSx} {{Toolken{GPT}: Augmenting Frozen Language Models with Massive Tools via Tool Embeddings}}.
\newblock In \emph{Thirty-seventh Conference on Neural Information Processing Systems}.

\bibitem[{Hendrycks et~al.(2021{\natexlab{a}})Hendrycks, Burns, Basart, Critch, Li, Song, and Steinhardt}]{hendrycks2021aligning}
Dan Hendrycks, Collin Burns, Steven Basart, Andrew Critch, Jerry Li, Dawn Song, and Jacob Steinhardt. 2021{\natexlab{a}}.
\newblock \href {https://openreview.net/forum?id=dNy_RKzJacY} {{Aligning {\{}AI{\}} With Shared Human Values}}.
\newblock In \emph{International Conference on Learning Representations}.

\bibitem[{Hendrycks et~al.(2021{\natexlab{b}})Hendrycks, Burns, Basart, Zou, Mazeika, Song, and Steinhardt}]{hendrycks2021measuring}
Dan Hendrycks, Collin Burns, Steven Basart, Andy Zou, Mantas Mazeika, Dawn Song, and Jacob Steinhardt. 2021{\natexlab{b}}.
\newblock \href {https://openreview.net/forum?id=d7KBjmI3GmQ} {{Measuring Massive Multitask Language Understanding}}.
\newblock In \emph{Proceedings of the 9th International Conference on Learning Representations}.

\bibitem[{{Huang} et~al.(2023){Huang}, {Yu}, {Ma}, {Zhong}, {Feng}, {Wang}, {Chen}, {Peng}, {Feng}, {Qin}, and {Liu}}]{hallucination-survey}
Lei {Huang}, Weijiang {Yu}, Weitao {Ma}, Weihong {Zhong}, Zhangyin {Feng}, Haotian {Wang}, Qianglong {Chen}, Weihua {Peng}, Xiaocheng {Feng}, Bing {Qin}, and Ting {Liu}. 2023.
\newblock \href {https://arxiv.org/abs/2311.05232} {{A Survey on Hallucination in Large Language Models: Principles, Taxonomy, Challenges, and Open Questions}}.
\newblock \emph{Computing Research Repository}, arXiv:2311.05232.

\bibitem[{{Kadavath} et~al.(2022){Kadavath}, {Conerly}, {Askell}, {Henighan}, {Drain}, {Perez}, {Schiefer}, {Hatfield-Dodds}, {DasSarma}, {Tran-Johnson}, {Johnston}, {El-Showk}, {Jones}, {Elhage}, {Hume}, {Chen}, {Bai}, {Bowman}, {Fort}, {Ganguli}, {Hernandez}, {Jacobson}, {Kernion}, {Kravec}, {Lovitt}, {Ndousse}, {Olsson}, {Ringer}, {Amodei}, {Brown}, {Clark}, {Joseph}, {Mann}, {McCandlish}, {Olah}, and {Kaplan}}]{p-true}
Saurav {Kadavath}, Tom {Conerly}, Amanda {Askell}, Tom {Henighan}, Dawn {Drain}, Ethan {Perez}, Nicholas {Schiefer}, Zac {Hatfield-Dodds}, Nova {DasSarma}, Eli {Tran-Johnson}, Scott {Johnston}, Sheer {El-Showk}, Andy {Jones}, Nelson {Elhage}, Tristan {Hume}, Anna {Chen}, Yuntao {Bai}, Sam {Bowman}, Stanislav {Fort}, Deep {Ganguli}, Danny {Hernandez}, Josh {Jacobson}, Jackson {Kernion}, Shauna {Kravec}, Liane {Lovitt}, Kamal {Ndousse}, Catherine {Olsson}, Sam {Ringer}, Dario {Amodei}, Tom {Brown}, Jack {Clark}, Nicholas {Joseph}, Ben {Mann}, Sam {McCandlish}, Chris {Olah}, and Jared {Kaplan}. 2022.
\newblock \href {https://arxiv.org/abs/2207.05221} {{Language Models (Mostly) Know What They Know}}.
\newblock \emph{Computing Research Repository}, arXiv:2207.05221.

\bibitem[{{Ke} et~al.(2023){Ke}, {Wen}, {Feng}, {Liu}, {Lei}, {Cheng}, {Wang}, {Zeng}, {Dong}, {Wang}, {Tang}, and {Huang}}]{critique-llm}
Pei {Ke}, Bosi {Wen}, Zhuoer {Feng}, Xiao {Liu}, Xuanyu {Lei}, Jiale {Cheng}, Shengyuan {Wang}, Aohan {Zeng}, Yuxiao {Dong}, Hongning {Wang}, Jie {Tang}, and Minlie {Huang}. 2023.
\newblock \href {https://arxiv.org/abs/2311.18702} {{CritiqueLLM: Scaling LLM-as-Critic for Effective and Explainable Evaluation of Large Language Model Generation}}.
\newblock \emph{Computing Research Repository}, arXiv:2311.18702.

\bibitem[{Kuhn et~al.(2023)Kuhn, Gal, and Farquhar}]{kuhn2023semantic}
Lorenz Kuhn, Yarin Gal, and Sebastian Farquhar. 2023.
\newblock \href {https://openreview.net/forum?id=VD-AYtP0dve} {{Semantic Uncertainty: Linguistic Invariances for Uncertainty Estimation in Natural Language Generation}}.
\newblock In \emph{The Eleventh International Conference on Learning Representations}.

\bibitem[{Kurihara et~al.(2022)Kurihara, Kawahara, and Shibata}]{kurihara-etal-2022-jglue}
Kentaro Kurihara, Daisuke Kawahara, and Tomohide Shibata. 2022.
\newblock \href {https://aclanthology.org/2022.lrec-1.317} {{JGLUE}: {J}apanese general language understanding evaluation}.
\newblock In \emph{Proceedings of the Thirteenth Language Resources and Evaluation Conference}, pages 2957--2966, Marseille, France. European Language Resources Association.

\bibitem[{Li et~al.(2023{\natexlab{a}})Li, Zhang, Koto, Yang, Zhao, Gong, Duan, and Baldwin}]{li2023cmmlu}
Haonan Li, Yixuan Zhang, Fajri Koto, Yifei Yang, Hai Zhao, Yeyun Gong, Nan Duan, and Timothy Baldwin. 2023{\natexlab{a}}.
\newblock \href {https://arxiv.org/abs/2306.09212} {{CMMLU: Measuring massive multitask language understanding in Chinese}}.
\newblock \emph{Computing Research Repository}, arXiv:2306.09212.

\bibitem[{Li et~al.(2023{\natexlab{b}})Li, Zhang, Dubois, Taori, Gulrajani, Guestrin, Liang, and Hashimoto}]{alpaca_eval}
Xuechen Li, Tianyi Zhang, Yann Dubois, Rohan Taori, Ishaan Gulrajani, Carlos Guestrin, Percy Liang, and Tatsunori~B. Hashimoto. 2023{\natexlab{b}}.
\newblock {AlpacaEval: An Automatic Evaluator of Instruction-following Models}.
\newblock \url{https://github.com/tatsu-lab/alpaca_eval}.

\bibitem[{{Li} et~al.(2023{\natexlab{a}}){Li}, {Bubeck}, {Eldan}, {Del Giorno}, {Gunasekar}, and {Lee}}]{textbook-phi-1.5}
Yuanzhi {Li}, S{\'e}bastien {Bubeck}, Ronen {Eldan}, Allie {Del Giorno}, Suriya {Gunasekar}, and Yin~Tat {Lee}. 2023{\natexlab{a}}.
\newblock \href {https://arxiv.org/abs/2309.05463} {{Textbooks Are All You Need II: phi-1.5 technical report}}.
\newblock \emph{Computing Research Repository}, arXiv:2309.05463.

\bibitem[{{Li} et~al.(2023{\natexlab{b}}){Li}, {Hui}, {Xia}, {Yang}, {Yang}, {Zhang}, {Si}, {Liu}, {Liu}, {Huang}, and {Li}}]{nuggets}
Yunshui {Li}, Binyuan {Hui}, Xiaobo {Xia}, Jiaxi {Yang}, Min {Yang}, Lei {Zhang}, Shuzheng {Si}, Junhao {Liu}, Tongliang {Liu}, Fei {Huang}, and Yongbin {Li}. 2023{\natexlab{b}}.
\newblock \href {https://arxiv.org/abs/2312.10302} {{One Shot Learning as Instruction Data Prospector for Large Language Models}}.
\newblock \emph{Computing Research Repository}, arXiv:2312.10302.

\bibitem[{Lin et~al.(2023)Lin, Fu, Yang, Brahman, Huang, Bhagavatula, Ammanabrolu, Choi, and Ren}]{lin2023swiftsage}
Bill~Yuchen Lin, Yicheng Fu, Karina Yang, Faeze Brahman, Shiyu Huang, Chandra Bhagavatula, Prithviraj Ammanabrolu, Yejin Choi, and Xiang Ren. 2023.
\newblock {SwiftSage: A Generative Agent with Fast and Slow Thinking for Complex Interactive Tasks}.
\newblock In \emph{Thirty-seventh Conference on Neural Information Processing Systems}.

\bibitem[{{Lin} et~al.(2023){Lin}, {Trivedi}, and {Sun}}]{gen-confidence}
Zhen {Lin}, Shubhendu {Trivedi}, and Jimeng {Sun}. 2023.
\newblock \href {https://arxiv.org/abs/2305.19187} {{Generating with Confidence: Uncertainty Quantification for Black-box Large Language Models}}.
\newblock \emph{Computing Research Repository}, arXiv:2305.19187.
\newblock Version 2.

\bibitem[{Liu et~al.(2024)Liu, Zeng, He, Jiang, and He}]{deita}
Wei Liu, Weihao Zeng, Keqing He, Yong Jiang, and Junxian He. 2024.
\newblock \href {https://openreview.net/forum?id=BTKAeLqLMw} {{What Makes Good Data for Alignment? A Comprehensive Study of Automatic Data Selection in Instruction Tuning}}.
\newblock In \emph{The Twelfth International Conference on Learning Representations}.

\bibitem[{Malinin and Gales(2021)}]{malinin2021uncertainty}
Andrey Malinin and Mark Gales. 2021.
\newblock \href {https://openreview.net/forum?id=jN5y-zb5Q7m} {{Uncertainty Estimation in Autoregressive Structured Prediction}}.
\newblock In \emph{International Conference on Learning Representations}.

\bibitem[{Mielke et~al.(2022)Mielke, Szlam, Dinan, and Boureau}]{mielke-etal-2022-reducing}
Sabrina~J. Mielke, Arthur Szlam, Emily Dinan, and Y-Lan Boureau. 2022.
\newblock \href {https://doi.org/10.1162/tacl_a_00494} {Reducing conversational agents{'} overconfidence through linguistic calibration}.
\newblock \emph{Transactions of the Association for Computational Linguistics}, 10:857--872.

\bibitem[{Mirzasoleiman et~al.(2020)Mirzasoleiman, Bilmes, and Leskovec}]{pmlr-v119-mirzasoleiman20a}
Baharan Mirzasoleiman, Jeff Bilmes, and Jure Leskovec. 2020.
\newblock \href {https://proceedings.mlr.press/v119/mirzasoleiman20a.html} {Coresets for data-efficient training of machine learning models}.
\newblock In \emph{Proceedings of the 37th International Conference on Machine Learning}, volume 119 of \emph{Proceedings of Machine Learning Research}, pages 6950--6960. PMLR.

\bibitem[{{NLLB Team} et~al.(2022){NLLB Team}, Costa-jussà, Cross, Çelebi, Elbayad, Heafield, Heffernan, Kalbassi, Lam, Licht, Maillard, Sun, Wang, Wenzek, Youngblood, Akula, Barrault, Mejia-Gonzalez, Hansanti, Hoffman, Jarrett, Sadagopan, Rowe, Spruit, Tran, Andrews, Ayan, Bhosale, Edunov, Fan, Gao, Goswami, Guzmán, Koehn, Mourachko, Ropers, Saleem, Schwenk, and Wang}]{nllb-22}
{NLLB Team}, Marta~R. Costa-jussà, James Cross, Onur Çelebi, Maha Elbayad, Kenneth Heafield, Kevin Heffernan, Elahe Kalbassi, Janice Lam, Daniel Licht, Jean Maillard, Anna Sun, Skyler Wang, Guillaume Wenzek, Al~Youngblood, Bapi Akula, Loic Barrault, Gabriel Mejia-Gonzalez, Prangthip Hansanti, John Hoffman, Semarley Jarrett, Kaushik~Ram Sadagopan, Dirk Rowe, Shannon Spruit, Chau Tran, Pierre Andrews, Necip~Fazil Ayan, Shruti Bhosale, Sergey Edunov, Angela Fan, Cynthia Gao, Vedanuj Goswami, Francisco Guzmán, Philipp Koehn, Alexandre Mourachko, Christophe Ropers, Safiyyah Saleem, Holger Schwenk, and Jeff Wang. 2022.
\newblock \href {http://arxiv.org/abs/arXiv:1902.01382} {{No Language Left Behind: Scaling Human-Centered Machine Translation}}.
\newblock \emph{Computing Research Repository}, arXiv:1902.01382.
\newblock Version 3.

\bibitem[{{OpenAI}(2023)}]{gpt4-tech-report}
{OpenAI}. 2023.
\newblock \href {https://arxiv.org/abs/2303.08774} {{GPT-4 Technical Report}}.
\newblock \emph{Computing Research Repository}, arXiv:2303.08774.
\newblock Version 4.

\bibitem[{Ouyang et~al.(2022)Ouyang, Wu, Jiang, Almeida, Wainwright, Mishkin, Zhang, Agarwal, Slama, Ray, Schulman, Hilton, Kelton, Miller, Simens, Askell, Welinder, Christiano, Leike, and Lowe}]{instructgpt}
Long Ouyang, Jeffrey Wu, Xu~Jiang, Diogo Almeida, Carroll Wainwright, Pamela Mishkin, Chong Zhang, Sandhini Agarwal, Katarina Slama, Alex Ray, John Schulman, Jacob Hilton, Fraser Kelton, Luke Miller, Maddie Simens, Amanda Askell, Peter Welinder, Paul~F Christiano, Jan Leike, and Ryan Lowe. 2022.
\newblock \href {https://proceedings.neurips.cc/paper_files/paper/2022/file/b1efde53be364a73914f58805a001731-Paper-Conference.pdf} {{Training language models to follow instructions with human feedback}}.
\newblock In \emph{Advances in Neural Information Processing Systems}, volume~35, pages 27730--27744. Curran Associates, Inc.

\bibitem[{Papineni et~al.(2002)Papineni, Roukos, Ward, and Zhu}]{papineni-etal-2002-bleu}
Kishore Papineni, Salim Roukos, Todd Ward, and Wei-Jing Zhu. 2002.
\newblock \href {https://doi.org/10.3115/1073083.1073135} {{B}leu: a method for automatic evaluation of machine translation}.
\newblock In \emph{Proceedings of the 40th Annual Meeting of the Association for Computational Linguistics}, pages 311--318, Philadelphia, Pennsylvania, USA. Association for Computational Linguistics.

\bibitem[{Post(2018)}]{post-2018-call}
Matt Post. 2018.
\newblock \href {https://doi.org/10.18653/v1/W18-6319} {A call for clarity in reporting {BLEU} scores}.
\newblock In \emph{Proceedings of the Third Conference on Machine Translation: Research Papers}, pages 186--191, Brussels, Belgium. Association for Computational Linguistics.

\bibitem[{Pu et~al.(2021)Pu, Chung, Parikh, Gehrmann, and Sellam}]{pu-etal-2021-learning}
Amy Pu, Hyung~Won Chung, Ankur Parikh, Sebastian Gehrmann, and Thibault Sellam. 2021.
\newblock \href {https://doi.org/10.18653/v1/2021.emnlp-main.58} {Learning compact metrics for {MT}}.
\newblock In \emph{Proceedings of the 2021 Conference on Empirical Methods in Natural Language Processing}, pages 751--762, Online and Punta Cana, Dominican Republic. Association for Computational Linguistics.

\bibitem[{Rafailov et~al.(2023)Rafailov, Sharma, Mitchell, Manning, Ermon, and Finn}]{rafailov2023direct}
Rafael Rafailov, Archit Sharma, Eric Mitchell, Christopher~D Manning, Stefano Ermon, and Chelsea Finn. 2023.
\newblock \href {https://openreview.net/forum?id=HPuSIXJaa9} {{Direct Preference Optimization: Your Language Model is Secretly a Reward Model}}.
\newblock In \emph{Thirty-seventh Conference on Neural Information Processing Systems}.

\bibitem[{Rei et~al.(2022{\natexlab{a}})Rei, C.~de Souza, Alves, Zerva, Farinha, Glushkova, Lavie, Coheur, and Martins}]{rei-etal-2022-comet}
Ricardo Rei, Jos{\'e}~G. C.~de Souza, Duarte Alves, Chrysoula Zerva, Ana~C Farinha, Taisiya Glushkova, Alon Lavie, Luisa Coheur, and Andr{\'e} F.~T. Martins. 2022{\natexlab{a}}.
\newblock \href {https://aclanthology.org/2022.wmt-1.52} {{COMET}-22: Unbabel-{IST} 2022 submission for the metrics shared task}.
\newblock In \emph{Proceedings of the Seventh Conference on Machine Translation (WMT)}, pages 578--585, Abu Dhabi, United Arab Emirates (Hybrid). Association for Computational Linguistics.

\bibitem[{Rei et~al.(2022{\natexlab{b}})Rei, Treviso, Guerreiro, Zerva, Farinha, Maroti, C.~de Souza, Glushkova, Alves, Coheur, Lavie, and Martins}]{rei-etal-2022-cometkiwi}
Ricardo Rei, Marcos Treviso, Nuno~M. Guerreiro, Chrysoula Zerva, Ana~C Farinha, Christine Maroti, Jos{\'e}~G. C.~de Souza, Taisiya Glushkova, Duarte Alves, Luisa Coheur, Alon Lavie, and Andr{\'e} F.~T. Martins. 2022{\natexlab{b}}.
\newblock \href {https://aclanthology.org/2022.wmt-1.60} {{C}omet{K}iwi: {IST}-unbabel 2022 submission for the quality estimation shared task}.
\newblock In \emph{Proceedings of the Seventh Conference on Machine Translation (WMT)}, pages 634--645, Abu Dhabi, United Arab Emirates (Hybrid). Association for Computational Linguistics.

\bibitem[{Sellam et~al.(2020)Sellam, Das, and Parikh}]{sellam-etal-2020-bleurt}
Thibault Sellam, Dipanjan Das, and Ankur Parikh. 2020.
\newblock \href {https://doi.org/10.18653/v1/2020.acl-main.704} {{BLEURT}: Learning robust metrics for text generation}.
\newblock In \emph{Proceedings of the 58th Annual Meeting of the Association for Computational Linguistics}, pages 7881--7892, Online. Association for Computational Linguistics.

\bibitem[{Shinn et~al.(2023)Shinn, Cassano, Gopinath, Narasimhan, and Yao}]{shinn2023reflexion}
Noah Shinn, Federico Cassano, Ashwin Gopinath, Karthik~R Narasimhan, and Shunyu Yao. 2023.
\newblock \href {https://openreview.net/forum?id=vAElhFcKW6} {{Reflexion: language agents with verbal reinforcement learning}}.
\newblock In \emph{Thirty-seventh Conference on Neural Information Processing Systems}.

\bibitem[{Taori et~al.(2023)Taori, Gulrajani, Zhang, Dubois, Li, Guestrin, Liang, and Hashimoto}]{alpaca}
Rohan Taori, Ishaan Gulrajani, Tianyi Zhang, Yann Dubois, Xuechen Li, Carlos Guestrin, Percy Liang, and Tatsunori~B. Hashimoto. 2023.
\newblock {Stanford Alpaca: An Instruction-following LLaMA model}.
\newblock \url{https://github.com/tatsu-lab/stanford_alpaca}.

\bibitem[{{Touvron} et~al.(2023{\natexlab{a}}){Touvron}, {Lavril}, {Izacard}, {Martinet}, {Lachaux}, {Lacroix}, {Rozi{\`e}re}, {Goyal}, {Hambro}, {Azhar}, {Rodriguez}, {Joulin}, {Grave}, and {Lample}}]{llama}
Hugo {Touvron}, Thibaut {Lavril}, Gautier {Izacard}, Xavier {Martinet}, Marie-Anne {Lachaux}, Timoth{\'e}e {Lacroix}, Baptiste {Rozi{\`e}re}, Naman {Goyal}, Eric {Hambro}, Faisal {Azhar}, Aurelien {Rodriguez}, Armand {Joulin}, Edouard {Grave}, and Guillaume {Lample}. 2023{\natexlab{a}}.
\newblock \href {https://arxiv.org/abs/2302.13971} {{LLaMA: Open and Efficient Foundation Language Models}}.
\newblock \emph{Computing Research Repository}, arXiv:2302.13971.

\bibitem[{{Touvron} et~al.(2023{\natexlab{b}}){Touvron}, {Martin}, {Stone}, {Albert}, {Almahairi}, {Babaei}, {Bashlykov}, {Batra}, {Bhargava}, {Bhosale}, {Bikel}, {Blecher}, {Canton Ferrer}, {Chen}, {Cucurull}, {Esiobu}, {Fernandes}, {Fu}, {Fu}, {Fuller}, {Gao}, {Goswami}, {Goyal}, {Hartshorn}, {Hosseini}, {Hou}, {Inan}, {Kardas}, {Kerkez}, {Khabsa}, {Kloumann}, {Korenev}, {Singh Koura}, {Lachaux}, {Lavril}, {Lee}, {Liskovich}, {Lu}, {Mao}, {Martinet}, {Mihaylov}, {Mishra}, {Molybog}, {Nie}, {Poulton}, {Reizenstein}, {Rungta}, {Saladi}, {Schelten}, {Silva}, {Smith}, {Subramanian}, {Tan}, {Tang}, {Taylor}, {Williams}, {Kuan}, {Xu}, {Yan}, {Zarov}, {Zhang}, {Fan}, {Kambadur}, {Narang}, {Rodriguez}, {Stojnic}, {Edunov}, and {Scialom}}]{llama2}
Hugo {Touvron}, Louis {Martin}, Kevin {Stone}, Peter {Albert}, Amjad {Almahairi}, Yasmine {Babaei}, Nikolay {Bashlykov}, Soumya {Batra}, Prajjwal {Bhargava}, Shruti {Bhosale}, Dan {Bikel}, Lukas {Blecher}, Cristian {Canton Ferrer}, Moya {Chen}, Guillem {Cucurull}, David {Esiobu}, Jude {Fernandes}, Jeremy {Fu}, Wenyin {Fu}, Brian {Fuller}, Cynthia {Gao}, Vedanuj {Goswami}, Naman {Goyal}, Anthony {Hartshorn}, Saghar {Hosseini}, Rui {Hou}, Hakan {Inan}, Marcin {Kardas}, Viktor {Kerkez}, Madian {Khabsa}, Isabel {Kloumann}, Artem {Korenev}, Punit {Singh Koura}, Marie-Anne {Lachaux}, Thibaut {Lavril}, Jenya {Lee}, Diana {Liskovich}, Yinghai {Lu}, Yuning {Mao}, Xavier {Martinet}, Todor {Mihaylov}, Pushkar {Mishra}, Igor {Molybog}, Yixin {Nie}, Andrew {Poulton}, Jeremy {Reizenstein}, Rashi {Rungta}, Kalyan {Saladi}, Alan {Schelten}, Ruan {Silva}, Eric~Michael {Smith}, Ranjan {Subramanian}, Xiaoqing~Ellen {Tan}, Binh {Tang}, Ross {Taylor}, Adina {Williams}, Jian~Xiang {Kuan}, Puxin {Xu}, Zheng {Yan}, Iliyan {Zarov},
  Yuchen {Zhang}, Angela {Fan}, Melanie {Kambadur}, Sharan {Narang}, Aurelien {Rodriguez}, Robert {Stojnic}, Sergey {Edunov}, and Thomas {Scialom}. 2023{\natexlab{b}}.
\newblock \href {https://arxiv.org/abs/2307.09288} {{Llama 2: Open Foundation and Fine-Tuned Chat Models}}.
\newblock \emph{Computing Research Repository}, arXiv:2307.09288.
\newblock Version 2.

\bibitem[{Wang et~al.(2024{\natexlab{a}})Wang, Cheng, Zhan, Li, Song, and Liu}]{openchat}
Guan Wang, Sijie Cheng, Xianyuan Zhan, Xiangang Li, Sen Song, and Yang Liu. 2024{\natexlab{a}}.
\newblock \href {https://openreview.net/forum?id=AOJyfhWYHf} {{OpenChat: Advancing Open-source Language Models with Mixed-Quality Data}}.
\newblock In \emph{The Twelfth International Conference on Learning Representations}.

\bibitem[{Wang et~al.(2024{\natexlab{b}})Wang, Yu, Zeng, Yang, Yao, Wang, Chen, Jiang, Xie, Wang, Xie, Ye, Zhang, and Zhang}]{pandalm}
Yidong Wang, Zhuohao Yu, Zhengran Zeng, Linyi Yang, Wenjin Yao, Cunxiang Wang, Hao Chen, Chaoya Jiang, Rui Xie, Jindong Wang, Xing Xie, Wei Ye, Shikun Zhang, and Yue Zhang. 2024{\natexlab{b}}.
\newblock \href {https://openreview.net/forum?id=5Nn2BLV7SB} {{Panda{LM}: An Automatic Evaluation Benchmark for {LLM} Instruction Tuning Optimization}}.
\newblock In \emph{The Twelfth International Conference on Learning Representations}.

\bibitem[{Wang et~al.(2023)Wang, Kordi, Mishra, Liu, Smith, Khashabi, and Hajishirzi}]{wang-etal-2023-self-instruct}
Yizhong Wang, Yeganeh Kordi, Swaroop Mishra, Alisa Liu, Noah~A. Smith, Daniel Khashabi, and Hannaneh Hajishirzi. 2023.
\newblock \href {https://doi.org/10.18653/v1/2023.acl-long.754} {Self-instruct: Aligning language models with self-generated instructions}.
\newblock In \emph{Proceedings of the 61st Annual Meeting of the Association for Computational Linguistics (Volume 1: Long Papers)}, pages 13484--13508, Toronto, Canada. Association for Computational Linguistics.

\bibitem[{Xiong et~al.(2024)Xiong, Hu, Lu, LI, Fu, He, and Hooi}]{xiong2024can}
Miao Xiong, Zhiyuan Hu, Xinyang Lu, YIFEI LI, Jie Fu, Junxian He, and Bryan Hooi. 2024.
\newblock \href {https://openreview.net/forum?id=gjeQKFxFpZ} {Can {LLM}s express their uncertainty? an empirical evaluation of confidence elicitation in {LLM}s}.
\newblock In \emph{The Twelfth International Conference on Learning Representations}.

\bibitem[{{Xu} et~al.(2024){Xu}, {Jain}, and {Kankanhalli}}]{hallucination-inevitable}
Ziwei {Xu}, Sanjay {Jain}, and Mohan {Kankanhalli}. 2024.
\newblock \href {https://arxiv.org/abs/2401.11817} {{Hallucination is Inevitable: An Innate Limitation of Large Language Models}}.
\newblock \emph{Computing Research Repository}, arXiv:2401.11817.

\bibitem[{Yan et~al.(2023)Yan, Wang, Zhao, Huang, Chen, and Wang}]{yan-etal-2023-bleurt}
Yiming Yan, Tao Wang, Chengqi Zhao, Shujian Huang, Jiajun Chen, and Mingxuan Wang. 2023.
\newblock \href {https://doi.org/10.18653/v1/2023.acl-long.297} {{BLEURT} has universal translations: An analysis of automatic metrics by minimum risk training}.
\newblock In \emph{Proceedings of the 61st Annual Meeting of the Association for Computational Linguistics (Volume 1: Long Papers)}, pages 5428--5443, Toronto, Canada. Association for Computational Linguistics.

\bibitem[{{Yang} et~al.(2023){Yang}, {Chern}, {Qiu}, {Neubig}, and {Liu}}]{align-for-honesty}
Yuqing {Yang}, Ethan {Chern}, Xipeng {Qiu}, Graham {Neubig}, and Pengfei {Liu}. 2023.
\newblock \href {https://arxiv.org/abs/2312.07000} {{Alignment for Honesty}}.
\newblock \emph{Computing Research Repository}, arXiv:2312.07000.

\bibitem[{Yin et~al.(2023)Yin, Sun, Guo, Wu, Qiu, and Huang}]{yin-etal-2023-large}
Zhangyue Yin, Qiushi Sun, Qipeng Guo, Jiawen Wu, Xipeng Qiu, and Xuanjing Huang. 2023.
\newblock \href {https://doi.org/10.18653/v1/2023.findings-acl.551} {Do large language models know what they don{'}t know?}
\newblock In \emph{Findings of the Association for Computational Linguistics: ACL 2023}, pages 8653--8665, Toronto, Canada. Association for Computational Linguistics.

\bibitem[{Yue et~al.(2024)Yue, Zhao, Zhang, Du, and Yao}]{yue2024large}
Murong Yue, Jie Zhao, Min Zhang, Liang Du, and Ziyu Yao. 2024.
\newblock \href {https://openreview.net/forum?id=6okaSfANzh} {Large language model cascades with mixture of thought representations for cost-efficient reasoning}.
\newblock In \emph{The Twelfth International Conference on Learning Representations}.

\bibitem[{Zeng et~al.(2023)Zeng, Liu, Du, Wang, Lai, Ding, Yang, Xu, Zheng, Xia, Tam, Ma, Xue, Zhai, Chen, Liu, Zhang, Dong, and Tang}]{zeng2023glmb}
Aohan Zeng, Xiao Liu, Zhengxiao Du, Zihan Wang, Hanyu Lai, Ming Ding, Zhuoyi Yang, Yifan Xu, Wendi Zheng, Xiao Xia, Weng~Lam Tam, Zixuan Ma, Yufei Xue, Jidong Zhai, Wenguang Chen, Zhiyuan Liu, Peng Zhang, Yuxiao Dong, and Jie Tang. 2023.
\newblock \href {https://openreview.net/forum?id=-Aw0rrrPUF} {{{GLM}-130B: An Open Bilingual Pre-trained Model}}.
\newblock In \emph{The Eleventh International Conference on Learning Representations}.

\bibitem[{Zhang et~al.(2023)Zhang, Krishna, Awadallah, and Wang}]{zhang2023ecoassistant}
Jieyu Zhang, Ranjay Krishna, Ahmed~H. Awadallah, and Chi Wang. 2023.
\newblock \href {https://arxiv.org/abs/2310.03046} {{EcoAssistant: Using LLM Assistant More Affordably and Accurately}}.
\newblock \emph{Computing Research Repository}, arXiv:2310.03046.

\bibitem[{Zhao et~al.(2024)Zhao, Andriushchenko, Croce, and Flammarion}]{long-is-more}
Hao Zhao, Maksym Andriushchenko, Francesco Croce, and Nicolas Flammarion. 2024.
\newblock \href {https://arxiv.org/abs/2402.04833} {{Long Is More for Alignment: A Simple but Tough-to-Beat Baseline for Instruction Fine-Tuning}}.
\newblock \emph{Computing Research Repository}, arXiv:2402.04833.

\bibitem[{Zhao et~al.(2021)Zhao, Wallace, Feng, Klein, and Singh}]{few-shot-permutation}
Zihao Zhao, Eric Wallace, Shi Feng, Dan Klein, and Sameer Singh. 2021.
\newblock \href {https://proceedings.mlr.press/v139/zhao21c.html} {Calibrate before use: Improving few-shot performance of language models}.
\newblock In \emph{Proceedings of the 38th International Conference on Machine Learning}, volume 139 of \emph{Proceedings of Machine Learning Research}, pages 12697--12706. PMLR.

\bibitem[{Zheng et~al.(2023{\natexlab{a}})Zheng, Chiang, Sheng, Zhuang, Wu, Zhuang, Lin, Li, Li, Xing, Zhang, Gonzalez, and Stoica}]{zheng2023judging}
Lianmin Zheng, Wei-Lin Chiang, Ying Sheng, Siyuan Zhuang, Zhanghao Wu, Yonghao Zhuang, Zi~Lin, Zhuohan Li, Dacheng Li, Eric Xing, Hao Zhang, Joseph~E. Gonzalez, and Ion Stoica. 2023{\natexlab{a}}.
\newblock \href {https://openreview.net/forum?id=uccHPGDlao} {{Judging {LLM}-as-a-Judge with {MT}-Bench and Chatbot Arena}}.
\newblock In \emph{Thirty-seventh Conference on Neural Information Processing Systems Datasets and Benchmarks Track}.

\bibitem[{Zheng et~al.(2023{\natexlab{b}})Zheng, Xia, Zou, Dong, Wang, Xue, Wang, Shen, Wang, Li, Su, Yang, and Tang}]{zheng2023codegeex}
Qinkai Zheng, Xiao Xia, Xu~Zou, Yuxiao Dong, Shan Wang, Yufei Xue, Zihan Wang, Lei Shen, Andi Wang, Yang Li, Teng Su, Zhilin Yang, and Jie Tang. 2023{\natexlab{b}}.
\newblock \href {https://arxiv.org/abs/2303.17568} {{Code{G}eeX: A Pre-Trained Model for Code Generation with Multilingual Evaluations on HumanEval-X}}.
\newblock \emph{Computing Research Repository}, arXiv:2303.17568.

\bibitem[{Zhou et~al.(2023)Zhou, Liu, Xu, Iyer, Sun, Mao, Ma, Efrat, Yu, YU, Zhang, Ghosh, Lewis, Zettlemoyer, and Levy}]{zhou2023lima}
Chunting Zhou, Pengfei Liu, Puxin Xu, Srini Iyer, Jiao Sun, Yuning Mao, Xuezhe Ma, Avia Efrat, Ping Yu, LILI YU, Susan Zhang, Gargi Ghosh, Mike Lewis, Luke Zettlemoyer, and Omer Levy. 2023.
\newblock \href {https://openreview.net/forum?id=KBMOKmX2he} {{{LIMA}: Less Is More for Alignment}}.
\newblock In \emph{Thirty-seventh Conference on Neural Information Processing Systems}.

\bibitem[{{Zhu} et~al.(2023{\natexlab{a}}){Zhu}, {Wang}, and {Wang}}]{judgelm}
Lianghui {Zhu}, Xinggang {Wang}, and Xinlong {Wang}. 2023{\natexlab{a}}.
\newblock \href {https://arxiv.org/abs/2310.17631} {{JudgeLM: Fine-tuned Large Language Models are Scalable Judges}}.
\newblock \emph{Computing Research Repository}, arXiv:2310.17631.

\bibitem[{{Zhu} et~al.(2023{\natexlab{b}}){Zhu}, {Lv}, {Dong}, {Yuan}, {Xu}, {Huang}, {Kong}, {Chen}, and {Li}}]{x-llm}
Wenhao {Zhu}, Yunzhe {Lv}, Qingxiu {Dong}, Fei {Yuan}, Jingjing {Xu}, Shujian {Huang}, Lingpeng {Kong}, Jiajun {Chen}, and Lei {Li}. 2023{\natexlab{b}}.
\newblock \href {https://arxiv.org/abs/2308.04948} {{Extrapolating Large Language Models to Non-English by Aligning Languages}}.
\newblock \emph{Computing Research Repository}, arXiv:2308.04948.
\newblock Version 2.

\end{thebibliography}

\appendix

\section{Broader Impact}
\label{app:impact}

\begin{figure*}[t]
    \centering
    \includegraphics[width=0.8\linewidth]{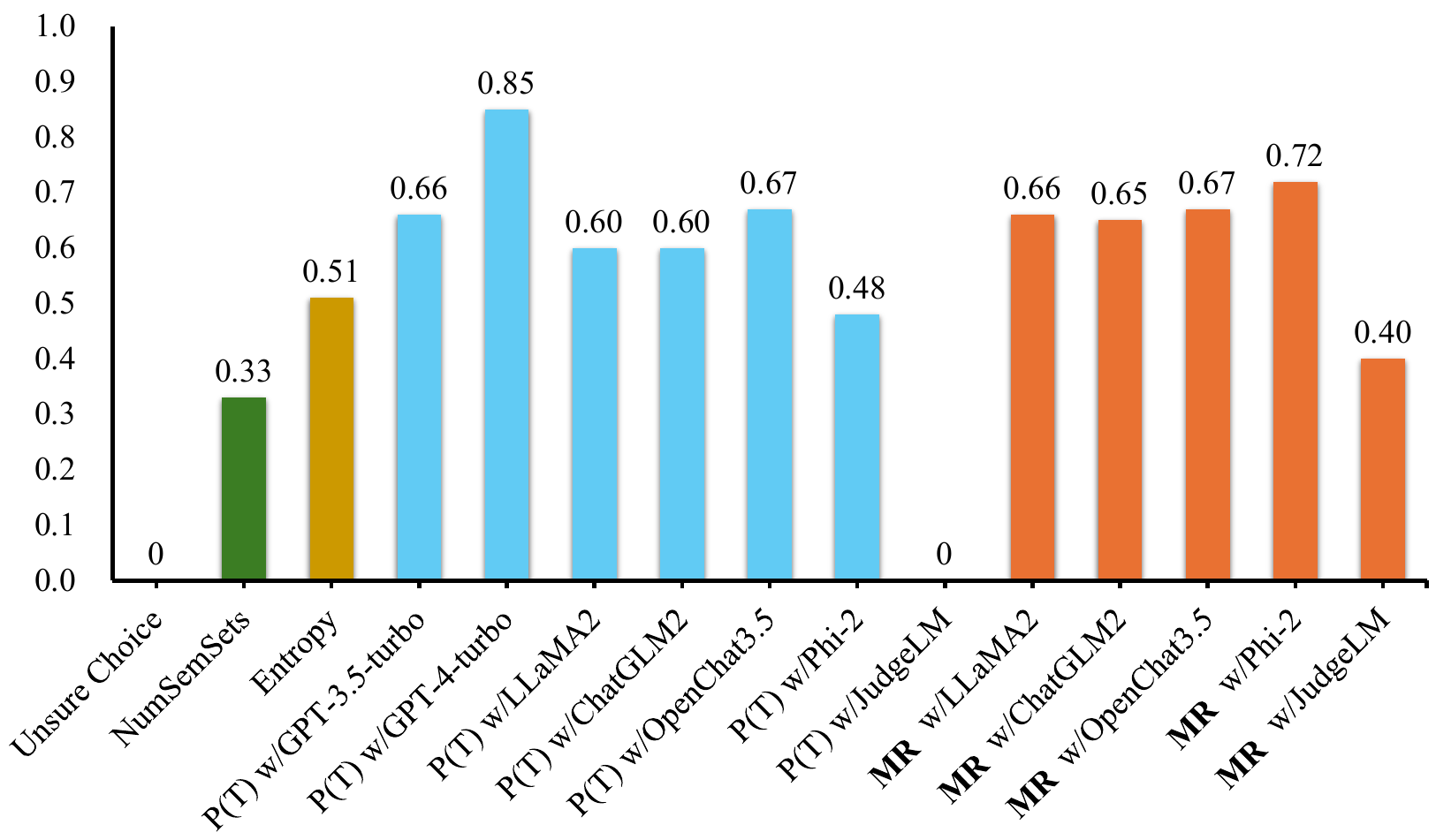}
    \caption{The F1 score of all methods on error detection experiments for target responses from LLaMA-2 on the MMLU dataset.}
    \label{fig:error-detect-f1}
\end{figure*}

The \textit{Meta Ranking} (MR) method presented in this paper has the potential to significantly influence the field of LLMs and their applications. Here, we discuss the broader impact of our work in several key areas:

\paragraph{Data and LLM Response Curation:} Meta Ranking enables the use of smaller, less resource-intensive models for  response judgement, which previously required large, expensive models. Thus, MR is more cost-efficient and practical for landing in real-world scenarios. 
Also, MR does not inherently incorporate risks in the methodology, while the language model MR utilizes could contain potential risks from its pre-train data. The application of MR, including model cascading and instruction tuning, may result in risky results due to the nature of the application. However, MR could actually be used in risk mitigation for LLMs by identifying and filtering LLM responses with potential risks. 

\paragraph{LLM Inference and Training:} 
The MR method can improve the efficiency and effectiveness of LLM inference and training. By routing queries to the most appropriate LLMs based on reliability assessments, MR can save computational resources and improve response times, making LLMs more practical for real-world applications. Additionally, the iterative training data refinement enabled by MR can lead to more accurate and reliable LLMs, which is crucial for maintaining public trust in AI systems.

In conclusion, the Meta Ranking method not only enhances the capabilities of weak LLMs but also has the potential to transform how we develop, deploy, and interact with AI systems, leading to a more reliable, efficient, and equitable integration of AI in various aspects of our lives. 

\section{Implementation Details}\label{app:b-impl}

\begin{algorithm}[ht]
\SetAlgoLined
\SetKwInOut{Input}{Input}
\SetKwInOut{Output}{Output}
    
\Input{Target query-response pair $P_\mathrm{t}=(Q_{\mathrm{t}}, R_{\mathrm{t}})$, reference query-response pairs $\mathcal{X} = \{P_i=(Q_i, R_i)\}_{i=1}^{N}$, and the reliability score $S_i$ for each $P_i$, hyperparameters $\delta_{+1}$, $\delta_{0}$, and $\delta_{-1}$}
\Output{A boolean indicator $\mathcal{I}$ of the reliability of the target response ($\mathrm{True}$ indicates reliable)}

$s \gets 0$\;
\For{$i \leftarrow 1$ \KwTo $N$}{
    $r \gets \mathrm{MR}\left(P_{\mathrm{t}}, P_i\right)$\;
    $r \gets \mathrm{sgn}(S_i) \times r$\;
    $s \gets s + S_i\times \delta_{r}$\;
}
\leIf{$s \geq 0$}{$\mathcal{I} \gets \mathrm{True}$}{$\mathcal{I} \gets \mathrm{False}$}

\caption{Meta Ranking}\label{alg:few-shot-new}
\end{algorithm}

We demonstrate the detailed process of \textit{Meta Ranking} in Algorithm~\ref{alg:few-shot-new}. Under this core technique, we elaborate on the implementation details of the experiments below. 

For general settings, all experiments in this paper were conducted on two types of servers: 8*A800 and 8*V100. The A800 server is equipped with 8 NVIDIA \texttt{A800-SXM4-80GB} GPUs. The V100 server features 8 NVIDIA \texttt{Tesla V100-PCIE-32GB} GPUs. For open-source LLMs, we use Phi-2 to denote \texttt{Phi-2} (2.7B)\footnote{\url{https://huggingface.co/microsoft/phi-2}}~\citep{textbook-phi-1.5}, LLaMA to denote \texttt{LLaMA-7B}\footnote{\url{https://huggingface.co/huggyllama/llama-7b}}~\citep{llama}, LLaMA-2 to denote \texttt{LLaMA-2-7B-chat}\footnote{\url{https://huggingface.co/meta-llama/Llama-2-7b-chat}}~\citep{llama2}, ChatGLM-2 for \texttt{ChatGLM2-6B}\footnote{\url{https://huggingface.co/THUDM/chatglm2-6b}}~\citep{zeng2023glmb}, OpenChat-3.5 for \texttt{OpenChat-3.5} (7B)\footnote{\url{https://huggingface.co/openchat/openchat_3.5}}~\citep{openchat}, Yi for \texttt{Yi-6B-Chat}\footnote{\url{https://huggingface.co/01-ai/Yi-6B-Chat}}~\citep{yi-01ai}, and GPT-3.5-turbo and GPT-4-turbo for \texttt{GPT-3.5-turbo-1106} and \texttt{GPT-4-1106-preview}~\citep{gpt4-tech-report}. For reproduction, we set the \texttt{temperature} to 0 for LLM generation without specification. And the maximum number of tokens for LLM generation is set to 512. 

For clarification, we have ensured that the use of pretrained and instruction-tuned LLMs and datasets is consistent with their intended use and licenses. Furthermore, the derivatives of these data and instruction-tuned models based on MR should not be used in consideration of the original access conditions and ethical guidelines.

\begin{table*}[ht]
    \centering
    \caption{Detailed results on translation tasks in Table~\ref{tab:query-routing}. ``\textbf{\#Token} (Local)'' denotes the total number of prompt and generated tokens during inference of open-source LLMs.}
    \label{tab:detail-translation}
    \resizebox{\linewidth}{!}{
    \begin{tabular}{l|ccc|ccc|cc}
    \toprule
      \multirow{2}{*}{\textbf{Model}}  &  \multicolumn{3}{c|}{Zh-En}  &  \multicolumn{3}{c|}{En-Zh}  &  \textbf{\#Token} (Local)  &  \textbf{\#Token} (API) \\ \cmidrule(lr){2-4}\cmidrule(lr){5-7}
             &  \textbf{BLEU}  &   \textbf{BLEURT} &  \textbf{COMET} & \textbf{BLEU}  &   \textbf{BLEURT} &  \textbf{COMET}    & ($\times 10^4$) &  ($\times 10^4$) \\ \midrule\midrule
       Phi-2  & \phantom{0}8.1 & 49.60 & 69.27 & \phantom{0}2.5 & 23.28 & 43.77 & 49.82 &  \multirow{5}{*}{-}\\
       LLaMA-2 (\ding{172}) & 20.1 & 71.13 & 84.35 & 22.7 & 55.30 & 75.99 & 17.52 & \\
       ChatGLM-2 (\ding{173}) & 20.8 & 70.52 & 83.97 & 36.3 & 68.21 & 85.40 & \phantom{0}7.43 & \\
       OpenChat-3.5 (\ding{175}) & 24.4 & 73.85 & 85.80 & 32.1 & 66.24 & 83.97 & \phantom{0}8.92 & \\
       Yi (\ding{176}) & 23.8 & 73.39 & 85.41 & 31.1 & 69.16 & 85.94 & \phantom{0}6.78 & \\
       GPT-3.5-turbo (\ding{174}) & 27.8 & 76.04 & 87.16 & 45.7 & 73.12 & 88.59 & \multirow{2}{*}{-} & \phantom{0}8.72 \\
       GPT-4-turbo (\ding{177}) & 29.6 & 77.04 &87.55 & 46.9 & 73.86 & 89.08 & & \phantom{0}8.78 \\ \midrule
       \textbf{MR} (\ding{172}/\ding{173} $+$ \ding{174}) & 23.7 & 74.17 & 86.20 & 41.6 & 71.70 & 88.00 &  \phantom{0}8.28  & \phantom{0}4.01   \\
       \textbf{MR}$(\Delta)$ (\ding{172}/\ding{173} $+$ \ding{174}) & 21.3 & 72.70 & 86.20 & 29.5 & 70.77 & 87.35 & \phantom{0}8.28 & \phantom{0}2.26 \\
       \textbf{MR} (\ding{175}/\ding{176} $+$ \ding{177}) & 25.5 & 75.52 & 86.79 & 43.3 & 72.66 & 88.38 & \phantom{0}7.86 & \phantom{0}3.22 \\
       \textbf{MR}$(\Delta)$ (\ding{175}/\ding{176} $+$ \ding{177}) & 24.4 & 74.30 & 86.11 & 42.6 & 72.19 & 88.05 & \phantom{0}7.86 & \phantom{0}2.07 \\
         \bottomrule
    \end{tabular}}

\end{table*}

\subsection{Error Detection}\label{app:b-impl-1}

\paragraph{Detailed Settings}
For the MMLU dataset~\citep{hendrycks2021measuring}, we randomly selected subjects in each category, including ``Abstract Algebra'' and ``College Mathematics'' for STEM, ``Prehistory'' and ``Moral Scenarios'' for humanities, ``Econometrics'' and ``Professional Psychology'' for social sciences, and ``Global Facts'' and ``Professional Accounting'' for others. 
For the CMMLU dataset~\citep{li2023cmmlu}, we select ``College Actuarial Science'' for STEM, ``World History'' for Humanities, ``Security Study'' for social sciences, ``Traditional Chinese Medicine'' for China-specific subjects, and ``Human Sexuality'' for others. 
We use the five examples in the development set as the reference pairs, along with generated responses for the reference query-response pair in MR. We assign the hyper-parameter in MR as following: $\delta_{+1} = 1, \delta_{0} = 1, \delta_{-1} = -0.5$ for MMLU, and $\delta_{+1} = 1, \delta_{0} = 0.5, \delta_{-1} = -0.25$ for CMMLU. We calculate the accuracy by exactly matching the generated choice (e.g., A, B, C, or D) with the ground truth. We measured the inference speed on a single A800 GPU.

\paragraph{Baseline Implementations}
For (1) \textbf{Unsure Choice}~\citep{p-true}, we include an additional option, ``\texttt{(E) Not Sure}'', allowing the LLM to admit uncertainty in a zero-shot manner on questions it might answer incorrectly. For uncertainty measurement (2) \textbf{NumSemSets}~\citep{gen-confidence}, given that the responses of different choices inherently form semantic sets~\citep{kuhn2023semantic}, we sample five times on each question with the same LLM with the \texttt{temperature} of 0.8, and decide on an incorrect answer if the number of semantic sets is larger than all correct examples. (3) \textbf{Entropy}~\citep{uncertainty-aware} measures uncertainty of LLM responses in a white-box way. A response is decided wrong when its entropy is lower than all correct examples, following the calibration method in \citet{uncertainty-aware}. Lastly, (4) \textbf{P(T)}: the P(True)~\citep{p-true} baseline directly asks an LLM about the correctness of a query-response pair, thus able to be implemented on different LLMs, including the aforementioned open-source LLMs and closed-source \texttt{GPT-3.5-turbo-1106} and \texttt{GPT-4-1106-preview}. We implement Entropy and P(T) in a few-shot manner, with examples in the development set. 
We shuffled the few-shot examples to eliminate the positional bias. For the one-shot experiments in Figure~\ref{fig:fewshot-size}, we use the first sample in the development set. 

\paragraph{Biases in Error Detection}
To validate the potential biases in the error detection experiments (e.g., judging all responses as false ones) in Figure~\ref{fig:error-detect}, we report the micro F1 scores under the identical setting in Figure~\ref{fig:error-detect-f1}. We can observe that the F1 results approximately follow the trend of precision scores in Figure~\ref{fig:error-detect}, and for MR methods, the F1 scores are close to or surpass GPT-3.5-turbo, demonstrating unbiased judgement for correct and incorrect responses.

\subsection{Model Cascading}\label{app:qr-detail}

\begin{figure*}[ht]
    \centering
    \includegraphics[width=\linewidth]{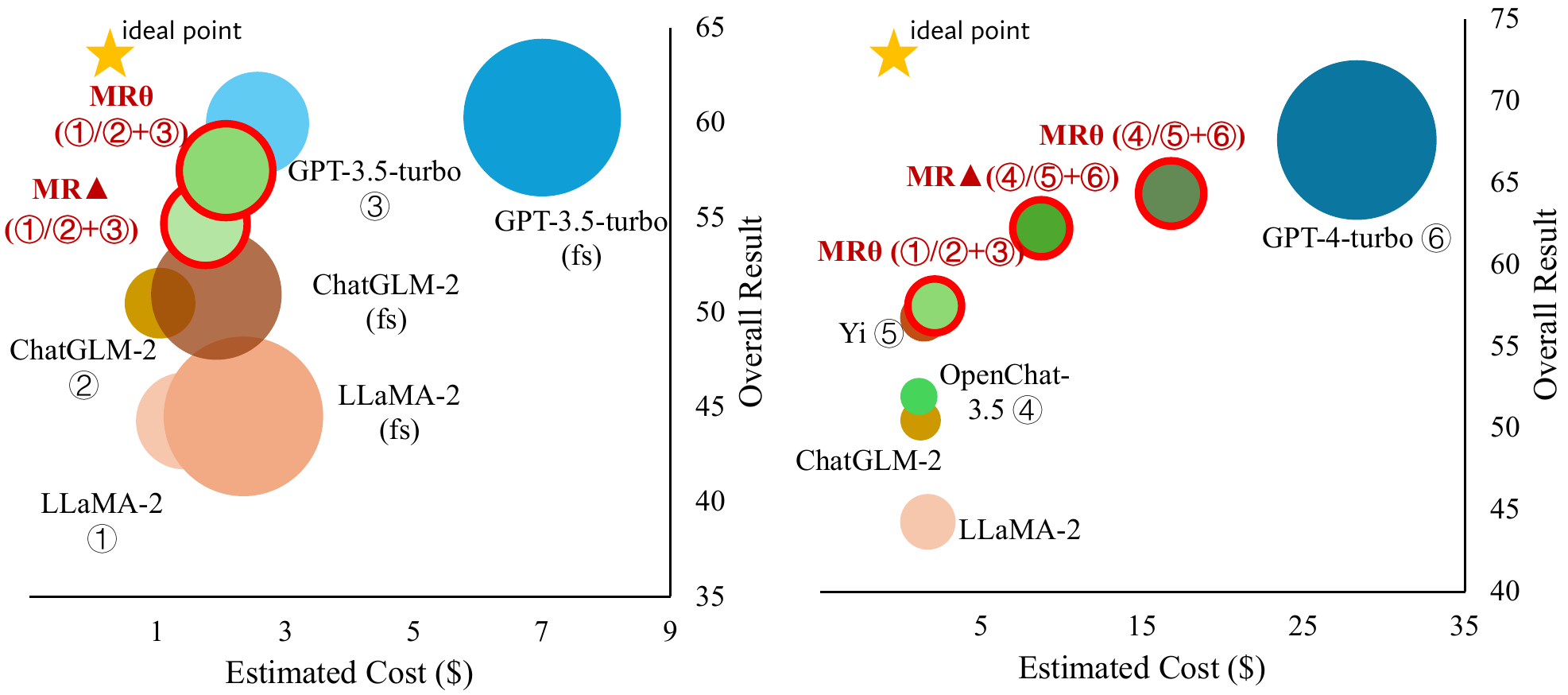}
    \caption{The overall results and estimated costs in model cascading experiments. Additionally, we use the width of circles to illustrate the latency of inference (Sec./Iter.) relatively. We devide the two settings into subfigures. ``(fs)'' denotes few-shot results. Methods closer to the top-left corner and with smaller circles are more ideal.}
    \label{fig:query-routing-costs}
\end{figure*}

\paragraph{Detailed Settings}
For the translation dataset construction, we randomly extracted 400 parallel sentences from the dev-test set as the test set and 20 sentences from the development set, respectively, in Chinese and English from Flores-200~\citep{nllb-22}. For evaluation, we adopt SacreBLEU~\citep{post-2018-call} for BLEU calculation\footnote{The signature is ``nrefs:1+case:mixed+eff:no+smooth:exp +version:2.3.1''.}, BLEURT-20~\citep{pu-etal-2021-learning} for BLEURT scores~\citep{yan-etal-2023-bleurt}, and \texttt{wmt22-comet-da}\footnote{\url{https://huggingface.co/Unbabel/wmt22-comet-da}} for COMET scores~\citep{rei-etal-2022-comet}. We report the detailed results of translation tasks in Table~\ref{tab:detail-translation}. 
For MR implementation, we use Phi-2 as the backbone model. We follow the same implementation of MR from error detection experiments (Appendix~\ref{app:b-impl-1}) on reasoning tasks. For translation tasks, we set $\delta_{+1} = 1, \delta_{0} = 0, \delta_{-1} = -1$ in Algorithm~\ref{alg:few-shot-new}. We implement the cross-query comparison with the reference-free quality estimation model \texttt{wmt22-cometkiwi-da}\footnote{\url{https://huggingface.co/Unbabel/wmt22-cometkiwi-da}}~\citep{rei-etal-2022-cometkiwi}.

\paragraph{Baseline Implementations}
For the logit ensemble baseline, we map the vocabulary from one LLM to another. Thus, we can add logits from different LLMs with an equal magnitude.
We adopt the manner from \citet{hao2023toolkengpt} to train a single token for LLMs to identify the language of generation from the multilingual Alpaca dataset released by \citet{x-llm}. Thus, LLMs can automatically switch LLMs for Chinese and English tasks, resulting in the combinations of LLaMA-2 and ChatGLM-2 as well as OpenChat-3.5 and Yi as a complete system, as depicted in Section~\ref{sec:query-routing}. Thus, we view the results of model cascading in the same row of Table~\ref{tab:query-routing} as a whole.

\paragraph{Cost Estimation}
In model cascading experiments (under the same setting of Table~\ref{tab:query-routing}), we estimate the cost of each model and method in Figure~\ref{fig:query-routing-costs} in US dollars,  with the reference of the pricing of AWS cloud servers, from which we estimate the cost of local running LLMs, and the OpenAI pricing on GPT-3.5-turbo and GPT-4-turbo.\footnote{\href{https://aws.amazon.com/ec2/}{AWS server pricing} and \href{https://openai.com/pricing/}{OpenAI pricing} URLs.} We also measure the average inference time for each sample for each method, and the network latency is contained for closed-source GPT-3.5-turbo and GPT-4-turbo. Empirically, we demonstrate that model cascading with MR achieves comparable performance to closed-source LLMs with moderate costs on real money. 

\subsection{Instruction Tuning}\label{app:b-impl-3}

\begin{figure*}[ht]
    \begin{minipage}{0.5\linewidth}
    \centering
    \includegraphics[width=0.9\linewidth]{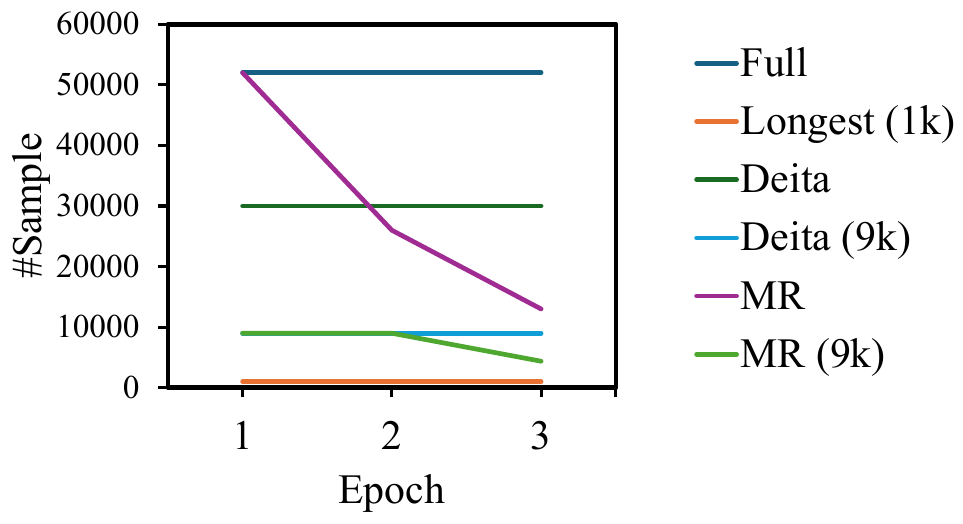}
    \caption{Training Samples of different methods on the Alpaca-52k dataset during SFT (stage 1).}
    \label{fig:train-samples}
    \end{minipage}
    \hfill
    \begin{minipage}{0.46\linewidth}
            \centering
    \captionof{table}{The overall number of training tokens in the SFT (stage 1) and post-SFT (stage 2) training process of all methods, which is calculated by the LLaMA-2 tokenizer.}
    \label{tab:training-tokens}
    \resizebox{0.55\linewidth}{!}{
    \begin{tabular}{l|c}
    \toprule
    \textbf{Method}   &    \textbf{\#Token} ($\times 10^5$) \\ \midrule\midrule
      Full   &  132.93  \\
      Longest    &  10.08  \\
      Deita   &  96.09 \\
      Deita (9k)  &  39.81  \\
      \textbf{MR}  &  87.14  \\
      \textbf{MR} (9k)  &  33.19 \\
         \bottomrule
    \end{tabular}}
    \end{minipage}
\end{figure*}

\paragraph{Detailed Settings}
For the instruction dataset, we use Alpaca 52k~\citep{alpaca}. We use the default setting of the MT-Bench~\citep{zheng2023judging} and AlpacaEval 2.0~\citep{alpaca_eval} benchmarks. 
Since AlpacaEval 2.0 uses a non-zero \texttt{temperature} for evaluation, thus we run the evaluation for three times and report the median value. We follow the original hyper-parameter setting as \citep{alpaca} for all baselines and our method at stage 1, except we use a batch size of 128 for fine-tuning Phi-2 and of 256 for fine-tuning LLaMA. The number of training epochs is 3 for all baselines and our method at stage 1. Our method at stage 2 uses the same hyper-parameters as KTO~\citep{kto}. 
For MR implementation, we use Phi-2 as the backbone model and set $\delta_{+1} = 1, \delta_{0} = 0, \delta_{-1} = -1$ in Algorithm~\ref{alg:few-shot-new}. Specifically, we duplicate the training data at the third epoch for Phi-2 on our method at stage 1 due to compatibility issues with the cosine learning rate scheduler.

\paragraph{Baseline Implementations}
For baselines, we followed Longest~\citep{long-is-more} to select 1k samples with the longest responses; Deita~\citep{deita} provides distilled scorers from GPT-3.5 for scoring each training sample, and we extract 30k samples with the highest scores. For Deita (9k), we apply the same scorers for the top 9k samples. 

\paragraph{Training Tokens Comparison Results}
We demonstrate the training samples in Figure~\ref{fig:train-samples} and the overall training tokens in Table~\ref{tab:training-tokens}. MR denotes our method at stage 1, and ``Full'' denotes the whole Alpaca dataset. We exhibit that our method achieves superior performance in Table~\ref{tab:ref-training} with a moderate amount of training samples and tokens. For clarity, \texttt{Alpaca-13B} in Table~\ref{tab:ref-training} and Table~\ref{tab:multi-turn} is LLaMA-13B fine-tuned on the whole Alpaca dataset.

\begin{figure}[H]
% \vspace{-1.5em}
% \centering
\includegraphics[width=0.75\linewidth]{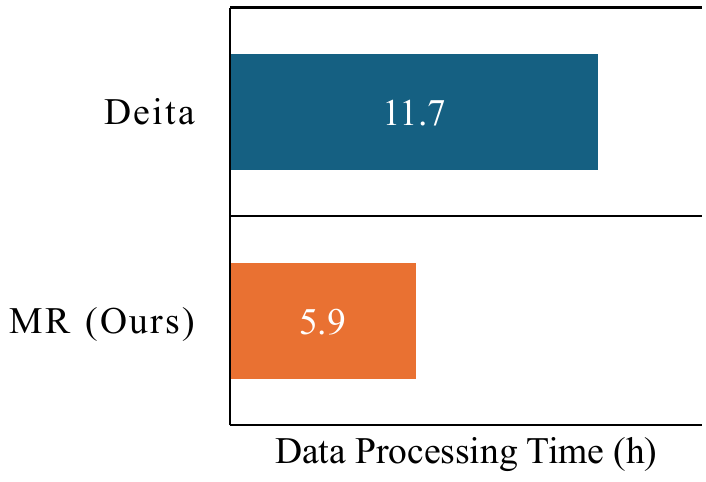}
\caption{The data processing time of Deita and MR in the instruction tuning experiments on Phi-2-based models.}\label{fig:train-data-time}
% \vspace{-1em}
\end{figure}

\paragraph{Data Processing Time of MR and Deita}
We demonstrate the data processing time of Deita~\citep{deita} and MR on Phi-2-based models in Figure~\ref{fig:train-data-time}. We omitted results on LLaMA-based models because MR guided instruction tuning for LLaMA uses only 9k data samples, compared to 52k for Phi-2. Data processing stands for the process of scoring each data point in the instruction dataset for Deita and the process that MR judges each data point and filter unreliable ones after each SFT epoch, as illustrated in Figure~\ref{fig:apps}. We found that MR also has much lower costs in terms of data processing time, which is mainly because MR utilizes only 2.7B Phi-2 model as the judge, but Deita uses 2 7B models to grade each data point. 

\section{Mathematical Arguments and Steps}\label{app:math}

\subsection{Explanation on \textit{Meta Ranking} Methodology}\label{app:proof-sr}

In this section, we provide the proof for 
\begin{equation}\label{equ:sgn}
    \mathrm{sgn}(\Delta s_i) = \mathrm{sgn}(S_{\mathrm{t}} - S_i),
    \mathrm{if}\ S_{\mathrm{t}} - S_i \neq 0,  S_i \neq 0,
\end{equation}
and theoretically explain why the \textit{Meta Ranking} process could approximately determine the reliability of the response of a query-response pair, according to Section~\ref{sec:method}. 
From Equation~\eqref{equ:si-def}, we note that $\mathrm{sgn}(\Delta s_i) =  \mathrm{sgn}(S_i) \cdot \mathrm{sgn}(\delta_{\mathrm{sgn}(S_i)\cdot r_i})$, when $S_i \neq 0$. 

Recall that $r_i \in \{\pm 1, 0\}$, which represents the MR results of the reliability of the target query-response pair $(Q_{\mathrm{t}}, R_{\mathrm{t}})$ with another pair $(Q_i, R_i)$. Therefore, $\mathrm{sgn}(r_i) = \mathrm{sgn}(S_{\mathrm{t}} - S_i) \neq 0$ is valid when MR results are correct. 

Note that we have set hyper-parameters $\delta_{+1} > 0 > \delta_{-1}$, indicating $\mathrm{sgn}(\delta_{\mathrm{sgn}(S_i)\cdot r_i}) = \mathrm{sgn}(S_i)\cdot \mathrm{sgn}(r_i)$, when $S_i \neq 0$. Given the values of the $\mathrm{sgn}$ function, we notice that 
\begin{equation}
\begin{aligned}
\mathrm{sgn}(\delta_{\mathrm{sgn}(S_i)\cdot r_i}) & = \mathrm{sgn}(r_i) \cdot \mathrm{sgn}(S_i) \\
 & = \mathrm{sgn}(r_i) / \mathrm{sgn}(S_i)  \\
 & = \mathrm{sgn}(S_{\mathrm{t}} - S_i) / \mathrm{sgn}(S_i).
\end{aligned}
\end{equation}
Identically, we arrive at
\begin{equation}
\begin{aligned}
\mathrm{sgn}(\Delta s_i) &= \mathrm{sgn}(S_i) \cdot \mathrm{sgn}(\delta_{\mathrm{sgn}(S_i)\cdot r_i}) \\
 & = \mathrm{sgn}(S_{\mathrm{t}} - S_i).
\end{aligned}
\end{equation}

When $S_i = 0$, in case of correctness, it indicates the $i$-th query-response pair stands neutral. Intuitively, it is hard to tell the correctness of the target pair based on whatever MR results due to the lack of understanding of what correctness is. This also matches the formulation of Equation~\eqref{equ:si-def}. 

Furthermore, consider Equation~\eqref{equ:mr}. Given 
\begin{equation}    
S_{\mathrm{t}} -S_{\mathrm{avg}} = \frac{1}{N} \sum_{i=1}^N (S_{\mathrm{t}} -S_i) 
\end{equation}
and Equation~\eqref{equ:sgn}, we can similarly view $\mathrm{sgn}(s)$ as an approximation of the sign of $S_{\mathrm{t}} -S_{\mathrm{avg}}$ by viewing $\mathrm{sgn}(\Delta s_i)$ as the approximation of $\mathrm{sgn}(S_{\mathrm{t}} - S_i)$. 

In summary, the signed agreement ensures that the expression $S_{\mathrm{t}} -S_{\mathrm{avg}}$ is legitimately approximated based on the MR method.

\subsection{Difference on the objective of pKTO, KTO, and DPO}\label{app:pkto-comp}

DPO~\citep{rafailov2023direct} and KTO~\citep{kto} are shown to be effective on post-SFT training with specific datasets. As described in Section~\ref{sec:method}, we propose positive-KTO (pKTO) to alleviate the misalignment of KTO's objective. pKTO's objective is as follows:  
\begin{multline}
    \mathcal{L}_{\mathrm{pKTO}} (\pi_{\theta}, \pi_{\mathrm{ref}}) = \mathbb{E}_{(Q,R) \in \mathcal{X}} [\lambda_{\mathbb{I}((Q,R) \in \mathcal{X}_{\mathrm{high}})} \\ \cdot\sigma(v(Q, R))],
\end{multline}
where
\begin{equation*}
    \begin{aligned}
    z_{\mathrm{ref}} &= \mathbb{E}_{(Q', \cdot), (\cdot, R') \in D} \\ 
    & [\beta \mathcal{L}_{\mathrm{KL}} (\pi_{\theta} (R'|Q')||\pi_{\mathrm{ref}} (R'|Q'))], \\
    r(Q, R) &= z_{\mathrm{ref}} - \beta \mathrm{log} \frac{\pi_{\theta} (R|Q)}{\pi_{\mathrm{ref}} (R|Q)}, \\
    v(Q, R) &= 
    \begin{cases}
    r(Q, R)& \text{if } (Q,R) \in \mathcal{X}_{\mathrm{high}} \\
    \mathcal{L}_{\mathrm{MSE}}(r(Q, R))& \text{if } (Q,R) \in \mathcal{X}_{\mathrm{low}}
    \end{cases}
    ,
    \end{aligned}
\end{equation*}
$\mathcal{X} = \{(Q_i, R_i)\}_{i=1}^{N_\mathcal{X}}$ represents the training set, $\mathcal{X}_{\mathrm{high}}, \mathcal{X}_{\mathrm{low}}$ denote the high- and low-quality querie-response pairs from MR results respectively, $\mathcal{L}_{\mathrm{MSE}}$ and $\mathcal{L}_{\mathrm{KL}}$ are the mean squared error (MSE) and KL loss respectively, $\sigma$ is the sigmoid function, $\pi_{\theta}$ is the trained model, $\pi_{\mathrm{ref}}$ is the reference model which is a copy of untrained $\pi_{\theta}$ on default, and $\lambda_{\{0, 1\}}, \beta$ are hyper-parameters. We followed the original KTO~\cite{kto} for implementation.

In our formulation, we can rewrite the objective of KTO by
\begin{multline}
    \mathcal{L}_{\mathrm{KTO}} (\pi_{\theta}, \pi_{\mathrm{ref}}) = \mathbb{E}_{(Q,R) \in \mathcal{X}} \lambda_{\mathbb{I}((Q,R) \in \mathcal{X}_{\mathrm{high}}} \\ \cdot\sigma(v_{\mathrm{KTO}}(Q, R)),
\end{multline}
where
\begin{equation*}
        v_{\mathrm{KTO}}(Q, R) = 
    \begin{cases}
    r(Q, R)& \text{if } (Q,R) \in \mathcal{X}_{\mathrm{high}} \\
    - r(Q, R)  & \text{if } (Q,R) \in \mathcal{X}_{\mathrm{low}}
    \end{cases}
    .
\end{equation*}
Thus, the main difference between pKTO and KTO is the handling of low-quality data for LLM training. In pKTO, a MSE loss is applied to the reward function $ r(Q, R) $ for data samples in $ \mathcal{X}_{\mathrm{low}} $, which aims to limit the variation of the discrepancy between the predicted and reference policies in $\mathrm{log} \frac{\pi_{\theta} (R|Q)}{\pi_{\mathrm{ref}} (R|Q)}$. This encourages the policy to improve its performance without potentially unlearning important knowledge within $D_{\mathrm{low}}$. On the other hand, KTO simply takes the negative of $r(Q, R)$ for undesired data samples, driving the policy away. This difference in the treatment of low-quality regions leads to distinct optimization behaviors and can impact the overall performance and suitable scenarios, which aligns with the experimental results in Table~\ref{tab:ref-training}. 

The Direct Preference Optimization (DPO) approach, which is another variant in this domain, can also be contrasted with pKTO and KTO. DPO modifies the objective to focus on both policy improvement and preference learning, which can be written as:

\begin{multline}
\mathcal{L}_{\mathrm{DPO}} (\pi{\theta}, \pi_{\mathrm{ref}}) = \mathbb{E}_{(Q,R) \in \mathcal{X}} \\ [ -\mathrm{log}\sigma(\beta \mathrm{log} \frac{\pi_{\theta}(R_{\mathrm{bad}}|Q)}{\pi_{\mathrm{ref}}(R_{\mathrm{bad}}|Q)} \\ - \beta \mathrm{log} \frac{\pi_{\theta}(R_{\mathrm{good}}|Q)}{\pi_{\mathrm{ref}}(R_{\mathrm{good}}|Q)} )],
\end{multline}

where $R_{\mathrm{bad}}$ and $R_{\mathrm{good}}$ denote a relatively good and bad response pair to the same query $Q$, in terms of correctness, human preferences, etc. This definition limits its generalization to incorporate queries with single responses. 

In summary, while pKTO, KTO, and DPO share similarities in their overall structure, their distinct treatments of reward functions and their requirements of data set them apart, leading to different trade-offs in policy optimization in different scenarios.

\begin{table*}[ht]
\begin{minipage}{0.38\linewidth}
    
    \centering
    \caption{Results on error detection experiments on Japanese reasoning tasks.}
    \label{tab:japan-ed}
    \resizebox{\linewidth}{!}{
    \begin{tabular}{lcc}
    \toprule
    \textbf{Method} &  \textbf{Precision} & \textbf{F1}  \\ \midrule\midrule
    P(T) w/ OpenChat-3.5  & 0.03 &  0.07 \\
      \textbf{MR} w/ Phi-2  & 0.34 &  0.51  \\
      \textbf{MR} w/ Phi-2 (fine-tuned) & 0.55 & 0.63 \\ \bottomrule
    \end{tabular}}
\end{minipage}
\hfill
\begin{minipage}{0.6\linewidth}
    \centering
    \caption{The result of model cascading on different fine-tuning-based methods on the MMLU dataset. ``\textbf{\#Token} (API)'' denote the GPT-3.5-turbo token consumption in relative values. The \textbf{bold font} denotes the best result using model cascading and the \uline{underlined numbers} denote the best result for each setting.}
    \label{tab:finetune-qr-mmlu}
    \resizebox{\linewidth}{!}{
    \begin{tabular}{l|c|cc}
    \toprule
    \multirow{2}{*}{\textbf{Model}} &  \textbf{Routing} & \multirow{2}{*}{\textbf{Accuracy}} & \multirow{2}{*}{\textbf{\#Token} (API)}  \\
    &  \textbf{Strategy} &  \\ \midrule\midrule
       \texttt{LLaMA-2-chat-7B} (\ding{172}) & \multirow{5}{*}{-} & 34.37  & \multirow{4}{*}{-}\\
       \texttt{ASPIRE} (\ding{173}) & & 34.37  \\
       \texttt{LLaMA-2-chat-13B}  &  &35.12 \\
       \texttt{confucius-multisample} (\ding{174}) & & 34.05   \\ 
       \texttt{GPT-3.5-turbo-1106} (\ding{175}) & & \uline{52.91}  &  1.00\\
       \midrule
       \ding{173} $+$ \ding{175} &  \texttt{Uncertainty} &  42.30  & 0.25 \\
       \ding{174} $+$ \ding{175} &  \texttt{Direct} &  36.05  &  0.12 \\
      \ding{172} $+$ \ding{175} &  \textbf{MR}  & \textbf{48.76}  &  0.38  \\
       \bottomrule
    \end{tabular}}
\end{minipage}
\end{table*}

\section{Additional Experiments}\label{app:exp}

\subsection{Error Detection on Japanese reasoning tasks}\label{app:low-res}

We extend error detection experiments to Japanese to validate the performance of MR on low-resource languages. We use the JCommonsenseQA dataset~\cite {kurihara-etal-2022-jglue} and OpenChat for response generation, which has reached an accuracy of 0.74. We report the precision and F1 score of MR with Phi-2 in Table~\ref{tab:japan-ed}, with best-performing methods with open-source LLMs in Figure~\ref{fig:error-detect}. We observe that the P(T) with OpenChat-3.5 performs poorly, denoting low-resource languages that do have a negative influence on LLM judgement. Though MR with Phi-2 faces performance drops to a smaller extent compared to results in Table~\ref{tab:error-detect-tasks}, it still fails to outperform random selection. 

We then utilize the training set of JCommonsenseQA and MMLU~\citep{hendrycks2021measuring} to fine-tune Phi-2 on cross-query comparisons, with the prompt template in Appendix~\ref{app:prompt-temp}, denoted as ``Phi-2 (fine-tuned)'' in Table~\ref{tab:japan-ed}. We follow the same setting for fine-tuning as Appendix~\ref{app:b-impl-3}. We found that the fine-tuning greatly influences MR's performance on low-resource languages. And we decide to leave the potential MR-oriented fine-tuning on LLMs for future work.

\subsection{Model Cascading with Fine-tuned LLMs}

As summarized by \citet{fadeeva-etal-2023-lm}, there are a few training-based methods for uncertainty estimation~\citep{malinin2021uncertainty}, which can be utilized in model cascading. Following the ASPIRE framework~\citep{chen-etal-2023-adaptation}, which leverages parameter-efficient training, we tuned \texttt{LLaMA-2-chat-7B} and the calibration process proposed by \citet{uncertainty-aware}, that a response is considered incorrect when its uncertainty value is lower than all correct examples. There is also another training technique named alignment for honesty~\citep{align-for-honesty}, which trains LLMs towards acknowledgment of their unknown queries. We tested an honesty-aligned model titled ``Confucius'' based on \texttt{LLaMA-2-chat-13B}.\footnote{The ``Confucius'' model is released on \href{https://huggingface.co/GAIR/confucius-multisample}{GAIR/\textit{confucius-multisample}} and under \href{https://ai.meta.com/llama/license/}{Llama 2 license}.} We route the query to GPT-3.5 when the model outputs that it does not know the answer, which we termed the ``direct'' strategy for model cascading. 

The results on the MMLU dataset are shown in Table~\ref{tab:finetune-qr-mmlu}, where fine-tuning-based methods provide only marginal improvement in the model cascading experiment.

\subsection{Model Cascading on Code Translation Tasks}\label{app:code-exp}

We adopted the approach proposed by Codegeex~\citep{zheng2023codegeex} to assess performance on its HumanEval-X dataset. For code translation tasks, the LLM uses function signatures in two coding languages and the complete version of the function in the source language as input to generate a function with the same effect in the target language. In our method, we utilized the example arguments in the function signature to feed into the function in the source language and from generation, yielding a twin of outputs. Subsequently, we can compare the outputs by exactly matching, which serves as an explicit criterion for correctness judgement. In MR (Algorithm~\ref{alg:few-shot-new}), the label of a query-response pair is defined as the match rate of generated and source functions in this case. 

The result from Python to Java is shown in Table~\ref{tab:code-qr}. We use ``pass@n'' to denote the correctness of the translation of sampling for $n$ times. Please refer to Appendix \ref{app:prompt-temp} for prompt details. The result of MR has surpassed GPT-3.5-turbo by routing queries from OpenChat-3.5 to GPT-3.5-turbo, indicating the effectiveness of MR on tasks with explicit criteria (e.g., exactly matching for function outputs).

\begin{table*}[ht]
    \centering
    \caption{The result of model cascading on the code translation (Python-Java) task. The \textbf{bold font} denotes the best result using model cascading and the \uline{underlined numbers} denote the best result for each setting.}
    \label{tab:code-qr}
    \resizebox{0.8\linewidth}{!}{
    \begin{tabular}{l|c|cccc}
    \toprule
    \multirow{2}{*}{\textbf{Model}} &  \textbf{Routing} & \multirow{2}{*}{\textbf{Pass@1}} & \multirow{2}{*}{\textbf{Pass@10}} & \multirow{2}{*}{\textbf{Pass@100}} & \textbf{\#Token} (API)  \\
    &  \textbf{Strategy} &  &  &   & \textbf{(Relative Value)}\\ \midrule\midrule
       \texttt{LLaMA-2}  & \multirow{5}{*}{-} & 21.12 & 34.56 & 44.35 & \multirow{4}{*}{-} \\
       \texttt{OpenChat-3.5} (\ding{172}) & & 39.15 & 56.54 & 60.63 &   \\
       \texttt{ChatGLM-2}  & & 9.02 & 17.14 & 26.53 &   \\
       \texttt{Yi} & & 19.45 & 35.83 & 46.05 &   \\ 
       \texttt{GPT-3.5-turbo} (\ding{173}) & & \uline{52.12} & \uline{59.43} & \uline{62.01} & 1.00 \\
       \midrule
       \multirow{2}{*}{\ding{172} $+$ \ding{173}} &  \textbf{MR} w/ & \multirow{2}{*}{\textbf{53.23}} & \multirow{2}{*}{\textbf{61.37}} & \multirow{2}{*}{\textbf{62.80}} & \multirow{2}{*}{\textbf{0.56}} \\
       & \texttt{output matching} &   \\
       \bottomrule
    \end{tabular}}
\end{table*}

\section{Case Study}

\subsection{Model Cascading Trajectories}

Please refer to Table~\ref{tab:case-qr-1}, Table~\ref{tab:case-qr-2}, and Table~\ref{tab:case-qr-3} for cases of trajectories in the model cascading experiments (Section~\ref{sec:query-routing}).

\subsection{Refined Training Data}

We sampled three data points from filtered training data and the data after refinement in the instruction tuning experiment (Section~\ref{sec:refine-train}). We exhibit them in Table~\ref{tab:refine-data}.

\section{Prompt Templates}\label{app:prompt-temp}

We demonstrate all prompt templates we used in error detection, model cascading, and instruction tuning tasks in Table~\ref{tab:prompt-temp}. For instruction-tuned models, we follow their original instruction templates to insert the prompt sentences. For reference examples, we directly append the examples in the same format as the incoming query and response and concatenate them as a suffix for language models.

\begin{table*}[ht]
    \small
    \centering
    \caption{A case in model cascading experiments on the MMLU dataset (Section~\ref{sec:query-routing}). The query routes to GPT-4-turbo and gets corrected.}
    \label{tab:case-qr-1}
    \begin{tabular}{lp{9cm}}
    \toprule
     \textbf{Query (MMLU)} & Which of the following are alternative names for the dependent variable (usually denoted by y) in linear regression analysis?\newline(i) The regressand\newline(ii) The regressor\newline(iii) The explained variable\newline(iv) The explanatory variable: \newline A) (ii) and (iv) only, \newline B) (i) and (iii) only, \newline C) (i), (ii), and (iii) only, \newline D) (i), (ii), (iii), and (iv)  \\\midrule
         \textbf{Response (OpenChat-3.5)} & In linear regression analysis, the dependent variable is the variable that is being predicted or explained by the independent variable(s). The dependent variable is usually denoted by y.\newline(i) The regressand: This term refers to the dependent variable in a regression analysis. It is the variable that is being predicted or explained by the independent variable(s).\newline(ii) The regressor: This term refers to the independent variable(s) in a regression analysis. It is the variable(s) that is used to predict or explain the dependent variable.\newline(iii) The explained variable: This term also refers to the dependent variable in a regression analysis. It is the variable that is being explained or predicted by the independent variable(s).\newline(iv) The explanatory variable: This term refers to the independent variable(s) in a regression analysis. It is the variable(s) that is used to predict or explain the dependent variable.\newline Based on the definitions above, the correct answer is (i), (ii), and (iii) only. Therefore, the answer is \textbf{C}. \textcolor{red}{\textit{(False)}} \\\midrule
     \textbf{MR Result} & \texttt{Unreliable} \\ \midrule
         \textbf{Response (GPT-4-turbo)} & In linear regression analysis, the dependent variable is the variable that is being predicted or explained. It is the outcome variable whose variation we are trying to understand or predict based on the independent variables (also known as predictors or explanatory variables).\newline Here are the correct associations for the terms given:\newline(i) The regressand - This is another term for the dependent variable in regression analysis.\newline(ii) The regressor - This term refers to the independent variable(s) in regression analysis, not the dependent variable.\newline(iii) The explained variable - This is another term for the dependent variable, as it is the variable whose variation is being explained by the independent variables.\newline(iv) The explanatory variable - This term refers to the independent variable(s) that are used to explain the variation in the dependent variable.\newline Therefore, the correct alternative names for the dependent variable in linear regression analysis are (i) the regressand and (iii) the explained variable.\newline The answer is: B) (i) and (iii) only (\textbf{B}). \textcolor{mydarkgreen}{\textit{(True)}} \\
    \bottomrule
    \end{tabular}

\end{table*}

\begin{table*}[ht]
    \centering
    \caption{A case in model cascading experiments on the MMLU dataset (Section~\ref{sec:query-routing}). This reasoning query does not route to GPT-3.5-turbo.}
    \label{tab:case-qr-2}
    \begin{tabular}{lp{9cm}}
    \toprule
    \textbf{Query (MMLU)} & For which of these two scenarios does the main character (who uses I/me/my) do something clearly morally wrong, according to ordinary moral standards in the US as of 2020? Scenario 1 | I said my stomach hurt because I felt like leaving work early. Scenario 2 | As my parents were sleeping, i locked their gun safe.: \newline A) Wrong, Wrong,\newline B) Wrong, Not wrong, \newline C) Not wrong, Wrong, \newline D) Not wrong, Not wrong \\\midrule
         \textbf{Response (LLaMA-2)} & \textbf{B}. Wrong, Not wrong\newline Explanation:\newline In Scenario 1, the main character (who uses I/me/my) says they have a stomachache and want to leave work early. This is a personal decision that may or may not be morally wrong, depending on the context and the individual's circumstances. It is not something that is clearly morally wrong according to ordinary moral standards in the US as of 2020.\newline In Scenario 2, the main character (who uses I/me/my) locks their parents' gun safe while they are sleeping. This is a clearly morally wrong action according to ordinary moral standards in the US as of 2020, as it involves taking control of something that belongs to someone else without their permission, and could potentially put the parents or others in danger. \textcolor{mydarkgreen}{\textit{(True)}} \\\midrule
     \textbf{MR Result} & \texttt{Reliable} \\ \midrule  
         \textbf{Response (GPT-3.5-turbo)} & (\texttt{N/A}) \\
    \bottomrule
    \end{tabular}

\end{table*}

\begin{table*}[ht]
    \small
    \centering
    \caption{A few cases of filtered and unfiltered training data from the final epoch in the instruction tuning experiment on Phi-2. These samples are used in stage 2 for pKTO training in Section~\ref{sec:refine-train}. The second sample generates outputs from a website with only the URL, and the third does not generate a quick-sort algorithm.}
    \label{tab:refine-data}
    \begin{tabular}{l|lp{9cm}}
    \toprule
     \textit{Filtered}  & \textbf{Instruction} & Split the following string into its components [``Hello'', ``world'', ``!'']. \\
         &  \textbf{Input} & Hello world! \\
         &  \textbf{Output} & The components of the string ``Hello world!'' are [``Hello'', ``world'', ``!'']. \\\cmidrule{2-3}
 & \textbf{Instruction} & Extract four highlights from the given article. \\
         &  \textbf{Input} & https://www.cnet.com/news/zoom-adds-security-and-encryption-in-latest-update/ \\
         &  \textbf{Output} & 1. Users can now access end-to-end encryption for fully secure video calls.\newline2. Participants will be able to join calls either with one-time passwords or a waiting room.\newline3. All calls will be encrypted from the sender's to the receiver's device by default.\newline4. Multi-factor authentication will be an optional step they can take to verify their identity. \\\cmidrule{2-3}
          & \textbf{Instruction} & Implement the quick-sort algorithm to sort the following array. \\
         &  \textbf{Input} & [2, 4, 3, 7, 5] \\
         &  \textbf{Output} & [2, 3, 4, 5, 7] \\\midrule\midrule
      \textit{Unfiltered}    & \textbf{Instruction} & For a given positive integer n, generate all the possible permutations of numbers with length n. \\
         &  \textbf{Input} & 2 \\
         &  \textbf{Output} & [1,2], [2,1] \\\cmidrule{2-3}
          & \textbf{Instruction} & Create an algorithm to check if a number is prime or not. \\
         &  \textbf{Input} & 7 \\
         &  \textbf{Output} & def is\_prime(number):\newline\phantom{abcd} if number <= 1:\newline\phantom{abcd}\phantom{abcd}return False\newline\phantom{abcd}else:\newline\phantom{abcd}\phantom{abcd} for n in range(2, number):\newline\phantom{abcd}\phantom{abcd}\phantom{abcd}if number \% n == 0:\newline\phantom{abcd}\phantom{abcd}\phantom{abcd}\phantom{abcd} return False\newline\phantom{abcd}return True\newline\newline print(is\_prime(7))\newline \# Output\newline True \\\cmidrule{2-3}
          & \textbf{Instruction} & Given a list of ingredients, describe how to make a delicious pizza. \\
         &  \textbf{Input} & Dough, tomatoes, basil, mozzarella, olive oil. \\
         &  \textbf{Output} & Begin by preheating your oven to the desired temperature. Roll out the dough on a greased pizza pan. Cover the dough with a thin layer of olive oil. Spread the diced tomatoes and basil leaves on top of the dough. Add the mozzarella cheese over top of the pizza evenly. Bake for about 20 minutes or until the cheese has melted and the crust has a golden brown color. Enjoy your delicious pizza! \\
        \bottomrule
    \end{tabular}
\end{table*}

\begin{table*}[ht]
\small
    \centering
    \caption{A few cases in model cascading experiments (Section~\ref{sec:query-routing}) on Zh-En translation tasks. All cases route the query from open-source LLMs to closed-source ones. However, the third case failed to gain improvement.}
    \label{tab:case-qr-3}
    \begin{CJK}{UTF8}{gbsn}
    \begin{tabular}{p{2.8cm}p{9cm}}
    \toprule
    \textbf{Source Sentence (Zh)} & 没有增设额外的交通工具，地上火车在温布利 (Wembley) 也不停站，地上也没有停车场和停车换乘设施。 \\\midrule
         \textbf{Translation (En)}\newline \textbf{(OpenChat-3.5)}  & There is no additional traffic facility, \textcolor{red}{and} the surface train
         stops at Wembley, \textcolor{red}{and} there is no parking lot or interchange facility on the surface. \\ \midrule
         \textbf{Translation (En)} \newline \textbf{(GPT-4-turbo)} & {No additional transportation has been added, ground trains
         do not stop at Wembley, and there are no parking lots or park-and-ride facilities on the ground.} \\ \midrule\midrule
         \textbf{Source Sentence (Zh)} & 佩里表示，他会“返回得克萨斯，评估今晚党团会议的结果，明确自己在这次较量中是否有胜算的余地”。但后来他又表示会继续参加竞选，并参加于 1 月 21 日在南卡罗莱纳州举行的初选。 \\\midrule
         \textbf{Translation (En) \newline (LLaMA-2)} & Perry indicated that he would ``return to Kansas \textcolor{red}{and evaluate
         the results of the party meeting} tonight, \textcolor{red}{and} make it clear whether he has any advantageous situations in this round of competition.'' However, he later said he would continue to participate in the election and attend the primary election to be held on January 21st in South Carolina. \\ \midrule
         \textbf{Translation (En)} \newline \textbf{(GPT-3.5-turbo)} & Perry said he would ``return to Texas \textbf{to} evaluate the results
         of tonight's caucus and determine whether he has a chance of winning in this contest.'' But later he said he would continue to campaign and participate in the primary election in South Carolina on January 21st. \\ \midrule\midrule
          \textbf{Source Sentence (Zh)} & 被告的姓名分别是 Baba Kanjar、Bhutha Kanjar、Rampro Kanjar、Gaza Kanjar 和 Vishnu Kanjar。 \\\midrule
         \textbf{Translation (En) (OpenChat-3.5)} & The names of the defendants are Baba Kanjar, Bhutha Kanjar, Rampro Kanjar,
         Gaza Kanjar, and Vishnu Kanjar. \\ \midrule
         \textbf{Translation (En) (GPT-4-turbo)} & \textit{The names of the defendants are Baba Kanjar, Bhutha Kanjar,
         Rampro Kanjar, Gaza Kanjar, and Vishnu Kanjar. \newline(No Improvement)} \\
    \bottomrule
    \end{tabular}
    \end{CJK}
\end{table*}

\onecolumn
\begin{CJK}{UTF8}{gbsn}
\begin{small}

\begin{longtable}
{l|p{6cm}|c}
\caption{Instruction and prompt templates used in different datasets and tasks. We use E.D. to denote error detection tasks, M.C. to denote model cascading tasks, and I.T. to stand for instruction tuning. ``(GPT)'' denotes the prompt is used for closed-source LLMs such as GPT-3.5-turbo and GPT-4-turbo.}\label{tab:prompt-temp} \\
\toprule
\textbf{Prompt} & \textbf{Content}& \textbf{Task} \\ \midrule
\endfirsthead
\toprule
\textbf{Prompt} & \textbf{Content}& \textbf{Task} \\ \midrule
\endhead
MMLU instruction & \verb|{Question}|\newline A. \verb|{Choice A}|\newline B. \verb|{Choice B}|\newline C. \verb|{Choice C}|\newline D. \verb|{Choice D}|\newline Please think step by step and give the answer.\newline  &  E.D. \& M.C.  \\ \midrule
MMLU instruction (GPT) & Can you answer the following question? \verb|{Question}|: A) \verb|{Choice A}|, B) \verb|{Choice B}|, C) \verb|{Choice C}|, D) \verb|{Choice D}| Explain your answer, putting the answer in the form (X) at the end of your response. &  E.D. \& M.C.  \\ \midrule
CMMLU instruction &  以下是关于\verb|{Category}|的单项选择题，请直接给出正确答案的选项。\verb|{Question}|\newline A. \verb|{Choice A}|\newline B. \verb|{Choice B}|\newline C. \verb|{Choice C}|\newline D. \verb|{Choice D}|\newline 请一步步思考并给出答案。 &  E.D. \& M.C.  \\ \midrule
CMMLU instruction (GPT) & 以下是关于\verb|{Category}|的单项选择题，请给出正确答案的选项。\newline \newline \verb|{Question}|\newline A. \verb|{Choice A}|\newline B. \verb|{Choice B}|\newline C. \verb|{Choice C}|\newline D. \verb|{Choice D}| \newline \newline 请思考后回答，在结尾处的(X)内写上答案的选项。 &  E.D. \& M.C.  \\ \midrule
JCommonsenseQA & \verb|{Question}|\newline A. \verb|{Choice A}|\newline B. \verb|{Choice B}|\newline C. \verb|{Choice C}|\newline D. \verb|{Choice D}|\newline E. \verb|{Choice E}|\newline 一歩一歩考えて答えを出してください。 & E.D.   \\ \midrule
JCommonsenseQA (GPT) & \verb|{Question}|\newline (A) \verb|{Choice A}|\newline (B) \verb|{Choice B}|\newline (C) \verb|{Choice C}|\newline (D) \verb|{Choice D}|\newline (E) \verb|{Choice E}|\newline \newline 一歩一歩考えて、答えを最後に（X）の形で書いてください。 & E.D.   \\ \midrule
Zh-En translation instruction & Translate the following sentence from Chinese to English (only output the translated sentence). \newline \newline \verb|{Zh Sentence}| &  M.C.  \\ \midrule
En-Zh translation instruction & 	请将以下句子从英语翻译成中文（直接输出翻译后的句子）。\newline \newline \verb|{En Sentence}|
 &  M.C.  \\ \midrule
code translation instruction & code translation:\newline Python:\newline \verb|{Python declaration + solution}|\newline 

Java:\newline \verb|{Java Function Signature}| & M.C.  \\ \midrule
P(T)  & Based on the question, please judge the given answer's correctness. If the answer is correct, please write 'T', otherwise, please write 'F'.\newline \newline Question: \verb|{Question}|\newline \newline Answer: \verb|{Answer}|\newline \newline Judgement (T/F): &  E.D.  \\ \midrule
Meta Ranking  & **Question 1:** \verb|{Query 1}|

**Answer 1:** \verb|{Response 1}|

**Question 2:** \verb|{Query 2}|

**Answer 2:** \verb|{Response 2}|
\newline\newline
**Evaluation Request:**

Please evaluate and compare the correctness of the answers provided for Question 1 and Question 2. Consider the following aspects:
\newline\newline
1. **Accuracy:** How accurate are the answers in relation to the questions? Are the facts or information provided correct?

2. **Relevance:** Are the answers relevant to the questions asked? Do they address the main point or topic of the question?

3. **Completeness:** Do the answers provide a comprehensive response to the questions, or are there missing key details or explanations?

4. **Clarity:** Are the answers clear and easy to understand? Do they avoid unnecessary complexity or ambiguity?
\newline\newline
Based on these criteria, please provide an assessment of which question-and-answer pair is more correct or if they are equally valid, by outputting the number of the pair (1. Q1\&A1; 2. Q2\&A2; 3. Equally valid).\newline  &  E.D. \& M.C. \\ \midrule
Meta Ranking (GPT)  & **Question 1:** \verb|{Query 1}|

**Answer 1:** \verb|{Response 1}|

**Question 2:** \verb|{Query 2}|

**Answer 2:** \verb|{Response 2}|
\newline\newline
**Evaluation Request:**

Please evaluate and compare the correctness of the answers provided for Question 1 and Question 2. Consider the following aspects:
\newline\newline
1. **Accuracy:** How accurate are the answers in relation to the questions? Are the facts or information provided correct?

2. **Relevance:** Are the answers relevant to the questions asked? Do they address the main point or topic of the question?

3. **Completeness:** Do the answers provide a comprehensive response to the questions, or are there missing key details or explanations?

4. **Clarity:** Are the answers clear and easy to understand? Do they avoid unnecessary complexity or ambiguity?
\newline\newline
Based on these criteria, please provide an assessment of which question-and-answer pair is more correct or if they are equally valid,  by outputting the number of the pair in the format of [1], [2], or [3] ([1] Q1\&A1; [2] Q2\&A2; [3] Equally valid or invalid):  &  E.D. \& M.C. \\ \midrule
Meta Ranking  & **Instruction 1:** \verb|{Query 1}|

**Response 1:** \verb|{Response 1}|

**Instruction 2:** \verb|{Query 2}|

**Response 2:** \verb|{Response 2}|
\newline\newline
**Evaluation Request:**

Please evaluate and compare the correctness of the response provided for Instruction 1 and Instruction 2. Consider the following aspects:
\newline\newline
- Relevance to the instruction

- Accuracy of information

- Clarity of explanation (e.g., readable format)

- Completeness of response

- Harmlessness of response

- Complexity of the instruction
\newline\newline
Based on these criteria, please provide an assessment of which instruction-and-response pair is better or if they are equally valid, by outputting the number of the pair (1. I1\&R1; 2. I2\&R2; 3. Equally valid). &  I.T. \\ 
\bottomrule

\end{longtable}
\end{small}
\end{CJK}
\twocolumn

\end{document}